\documentclass[runningheads]{llncs}
\usepackage[T1]{fontenc}
\usepackage[hidelinks]{hyperref}

\usepackage{graphicx}
\usepackage{color}

\hypersetup{
  colorlinks=true,
  linkcolor=black,
  urlcolor=blue,
  citecolor=black,
}
\usepackage{mwe}
\usepackage{acronym}
\usepackage{amsmath}
\usepackage{amssymb}
\usepackage{enumitem}
\usepackage{multirow}
\usepackage{tabularx}
\usepackage{booktabs}
\usepackage{subcaption}
\usepackage{threeparttable}
\newcommand{\Pxy}{\mathbb{P}_{XY}}
\newcommand{\Pmodel}{\mathbb{P}_F}
\newcommand{\ftrue}{f}
\newcommand{\fhat}{\hat f}

\newcommand{\X}{X}
\newcommand{\Y}{Y}
\newcommand{\xvec}{\mathbf x}
\newcommand{\Dataset}{\mathcal{D}}
\newcommand{\Dtrain}{\Dataset_{\text{train}}}
\newcommand{\Dtest}{\Dataset_{\text{test}}}
\newcommand{\Dval}{\Dataset_{\text{val}}}
\newcommand{\DatasetRV}{D}
\newcommand{\nsamples}{n}

\newcommand{\RiskL}{R_L}
\newcommand{\RiskEmp}{R_{\text{emp}}}

\newcommand{\E}{\mathbb{E}}
\newcommand{\Var}{\mathrm{Var}}

\newcommand{\pfeat}{p}
\newcommand{\Sidx}{S}
\newcommand{\notSidx}{\bar{S}} % \backslash S
\newcommand{\Sset}{\{1,\ldots,\pfeat\}}

\newcommand{\XS}{X_{\Sidx}}
\newcommand{\XnotS}{X_{\notSidx}}
\newcommand{\xS}{x_{\Sidx}}
\newcommand{\xnotS}{\mathbf{x}_{\notSidx}}

\newcommand{\func}{h}
\newcommand{\Xspace}{\mathcal{X}}
\newcommand{\Yspace}{\mathcal{Y}}

\newcommand{\PDsTheo}[1]{PD_{#1,\Sidx}}
\newcommand{\PDsEst}[1]{\widehat{PD}_{#1,\Sidx}}
\newcommand{\PDTheo}[1]{PD_{#1}}
\newcommand{\PDEst}[1]{\widehat{PD}_{#1}}
\newcommand{\cALEsTheo}[1]{ALE_{#1,\Sidx}}
\newcommand{\cALEsEst}[1]{\widehat{ALE}_{#1,\Sidx}}
\newcommand{\cALETheo}[1]{ALE_{#1}}
\newcommand{\cALEEst}[1]{\widehat{ALE}_{#1}}
\newcommand{\uALEsTheo}[1]{\widetilde{ALE}_{#1,\Sidx}}
\newcommand{\uALEsEst}[1]{\widehat{\widetilde{ALE}}_{#1,\Sidx}}
\newcommand{\uALETheo}[1]{\widetilde{ALE}_{#1}}
\newcommand{\uALEEst}[1]{\widehat{\widetilde{ALE}}_{#1}}
\newcommand{\uALETheoBinned}[1]{\widetilde{ALE}_{#1}^{\Kbins}}
\newcommand{\FinDiffSK}[1]{\Delta_{#1,\Sidx}^{\Kbins}}
\newcommand{\FinDiff}[1]{\Delta_{#1}}

\newcommand{\Kbins}{K}
\newcommand{\kbin}{k}
\newcommand{\IsKk}{I_{\Sidx}^{\Kbins}(\kbin)}
\newcommand{\Isk}{I_{\Sidx}(\kbin)}
\newcommand{\EventInI}{\mathcal{I}_k}
\newcommand{\zsKupper}{z_{\kbin,\Sidx}^{\Kbins}}
\newcommand{\zsKlower}{z_{\kbin-1,\Sidx}^{\Kbins}}

\newcommand{\nsKk}{\nsamples_{\Sidx}^{\Kbins}(\kbin)}
\newcommand{\nsk}{\nsamples_{\Sidx}(\kbin)}
\newcommand{\ksKx}{k_{\Sidx}^{\Kbins}(\xS)}
\newcommand{\ksx}{k_{\Sidx}(\xS)}
\newcommand{\BBinIndex}{B}
\acrodef{PD}{partial dependence}
\acrodef{ALE}{accumulated local effects}
\acrodef{XAI}{eXplainable AI}
\acrodef{IML}{interpretable machine learning}
\acrodef{PFI}{permutation feature importance}
\acrodef{ICE}{Individual conditional expectation}
\acrodef{MSE}{mean squared error}
\acrodef{MDI}{mean decrease in impurity}
\acrodef{GAM}{generalized additive model}
\acrodef{SVM}{support vector machine}
\acrodef{CV}{cross-validation}
\spnewtheorem*{proofsketch}{Proof sketch}{\itshape}{\rmfamily}

\begin{document}
\title{Analyzing Error Sources in Global Feature Effect Estimation}
%
% \titlerunning{}
\author{Timo Heiß\inst{1,2}\orcidID{0009-0002-0392-4308}
\and
Coco Bögel\inst{1,2}\orcidID{0009-0002-0683-4957}
\and
Bernd Bischl\inst{1,2}\orcidID{0000-0001-6002-6980}
\and
Giuseppe Casalicchio\inst{1,2}\orcidID{0000-0001-5324-5966}
}
\authorrunning{T. Heiß et al.}
\institute{LMU Munich, Munich, Germany
\email{\{timo.heiss,giuseppe.casalicchio\}@stat.uni-muenchen.de} \and
Munich Center for Machine Learning (MCML)
}
\maketitle              % typeset the header of the contribution
\begin{abstract}
    Global feature effects such as \ac{PD} and \ac{ALE} plots are widely used to interpret black-box models. However, they are only estimates of true underlying effects, and their reliability depends on multiple sources of error. 
    Despite the popularity of global feature effects, these error sources are largely unexplored. In particular, the practically relevant question of whether to use training or holdout data to estimate feature effects remains unanswered.
    We address this gap by providing a systematic, estimator-level analysis that disentangles sources of bias and variance for \ac{PD} and \ac{ALE}.
    To this end, we derive a mean-squared-error decomposition that separates model bias, estimation bias, model variance, and estimation variance, and analyze their dependence on model characteristics, data selection, and sample size.
    We validate our theoretical findings through an extensive simulation study across multiple data-generating processes, learners, estimation strategies (training data, validation data, and cross-validation), and sample sizes.
    Our results reveal that, while using holdout data is theoretically the cleanest, potential biases arising from the training data are empirically negligible and dominated by the impact of the usually higher sample size. 
    The estimation variance depends on both the presence of interactions and the sample size, with \ac{ALE} being particularly sensitive to the latter. 
    Cross-validation-based estimation is a promising approach that reduces the model variance component, particularly for overfitting models.
    Our analysis provides a principled explanation of the sources of error in feature effect estimates and offers concrete guidance on choosing estimation strategies when interpreting machine learning models.

\end{abstract}
%
%
%
%%%%%%%%%%%%%%%%%%%%%%%%%%%%%%%%%%%%%%%%%%%%%%%%%%%%%%%%%%%%%%%%%%
%
%                             INTRO
%
%%%%%%%%%%%%%%%%%%%%%%%%%%%%%%%%%%%%%%%%%%%%%%%%%%%%%%%%%%%%%%%%%%
\section{Introduction}

Many machine learning models are black boxes whose internal structure is not, or only partly, compatible with human reasoning, complicating explanations of both individual predictions and overall model behavior. This lack of transparency is especially problematic in high-stakes domains such as healthcare, law, and finance, where decisions must be transparent.
To address this challenge, the field of \ac{XAI} has proposed a wide range of methods to explain machine learning models~\cite{molnar_interpretable_2022}.
However, these methods must be used correctly to avoid misleading conclusions, as there are many pitfalls to be aware of~\cite{molnar_general_2022}.

Global feature effect methods such as \acf{PD}~\cite{friedman_greedy_2001} and \acf{ALE}~\cite{apley_visualizing_2020} visualize how one or more features affect predictions. %, marginalizing or locally integrating over the remaining features.
In practice, they are estimated from finite data, and their reliability depends on various error sources. 
Despite their widespread adoption, the error components of feature effect estimates remain largely unexplored.
Prior works focus on extrapolation under feature dependence~\cite{apley_visualizing_2020,gkolemis_rhale_2023}, aggregation bias~\cite{gkolemis_rhale_2023,goldstein_peeking_2015,herbinger_repid_2022,herbinger_decomposing_2024}, or quantifying uncertainty~\cite{apley_visualizing_2020,cook_explaining_2024,moosbauer_explaining_2021}. 
A formal bias–variance decomposition w.r.t. a true underlying effect has only been derived for the theoretical \ac{PD}~\cite{molnar_relating_2023}, leaving estimator-level errors introduced by finite data largely unaddressed.

A related practical question is whether to estimate explanations using training or holdout data. 
This has been studied for methods like \ac{PFI}~\cite{molnar_general_2022}, \ac{MDI}, and SHAP~\cite{loecher_debiasing_2022}, but remains open for \ac{PD} and \ac{ALE}.
While most works compute feature effects on training data~\cite{apley_visualizing_2020,friedman_greedy_2001,greenwell_pdp_2017,molnar_interpretable_2022}, other works use holdout data~\cite{molnar_relating_2023}. 
Practitioners still debate whether to estimate \ac{PD} and \ac{ALE} on training or holdout data (see \S\ref{sec:related-work}), trading larger training sample sizes against potential overfitting bias. These open questions point to a lack of an estimator-level understanding of error in global feature effect estimation, which our work addresses. 
Our main contributions are: 
\begin{itemize}[topsep=0pt]
    \item We provide the first estimator-level analysis of \ac{PD} and \ac{ALE}, deriving a full \acf{MSE} decomposition that separates model bias, estimation bias, model variance, and estimation variance.
    \item We theoretically analyze these components, showing how sample size and interactions affect estimation bias and variance differently for \ac{PD} and \ac{ALE}, and formally relating remaining bias and variance to model bias and variance.
    \item We empirically validate our findings in an extensive simulation study across multiple data-generating processes, learners, sample sizes, and estimation strategies (training, validation, and \ac{CV}), using dedicated estimators for the error components. We find negligible bias differences between training and holdout data, a strong sample-size effect -- especially for \ac{ALE} -- and that \ac{CV} is often preferable due to variance reduction.
\end{itemize}

%
%
%
%%%%%%%%%%%%%%%%%%%%%%%%%%%%%%%%%%%%%%%%%%%%%%%%%%%%%%%%%%%%%%%%%%
%
%                         Related Work
%
%%%%%%%%%%%%%%%%%%%%%%%%%%%%%%%%%%%%%%%%%%%%%%%%%%%%%%%%%%%%%%%%%%
\section{Related Work}\label{sec:related-work}

Many issues in feature effects have been analyzed. The \ac{PD} plot~\cite{friedman_greedy_2001} is known to suffer from extrapolation under dependent features by evaluating the model on implausible feature combinations~\cite{molnar_general_2022}.
\ac{ALE} plots avoid exactly that issue~\cite{apley_visualizing_2020}.
Another issue is \textit{aggregation bias}: global effects can obscure interaction-induced heterogeneity. \ac{ICE} curves~\cite{goldstein_peeking_2015} and RHALE~\cite{gkolemis_rhale_2023} visualize this heterogeneity, while regional methods like REPID~\cite{herbinger_repid_2022} and GADGET~\cite{herbinger_decomposing_2024} report effects in regions with reduced heterogeneity.
Several works quantify uncertainty via variance of feature effects with model-specific \ac{PD} confidence bands existing for probabilistic models~\cite{moosbauer_explaining_2021}, as well as model-agnostic approaches that consider \ac{PD} or \ac{ALE} across multiple model fits~\cite{cook_explaining_2024,molnar_relating_2023,apley_visualizing_2020}.
Only a few works explicitly study bias–variance trade-offs in feature effect estimators. 
RHALE~\cite{gkolemis_rhale_2023} optimizes \ac{ALE} binning to balance bias and variance but does not relate this trade-off to a ground-truth effect.
In~\cite{gkolemis_dale_2023}, bias and variance of the proposed DALE estimator w.r.t. ALE are analyzed.
In~\cite{chang_accelerated_2025}, variance reduction and consistency of their proposed ALE estimator A2D2E are shown, and the error of \ac{PD}, \ac{ALE}, and A2D2E estimators against a derivative-based ground-truth are compared in simulations.
Moreover, \cite{molnar_relating_2023} formalize \ac{PD} as a statistical estimator of a target estimand, derive a formal \ac{MSE} decomposition for the theoretical \ac{PD}, and propose variance estimators for confidence intervals.
In contrast, we provide the first full estimator-level \ac{MSE} decomposition of empirical \ac{PD} and \ac{ALE} w.r.t. their corresponding ground-truth feature effects, and analyze the error and its components both theoretically and empirically.

For many interpretability methods, computing explanations on training vs. holdout data can affect conclusions: it matters for loss-based methods like \acf{PFI}~\cite{molnar_general_2022} and for \acf{MDI} and SHAP, which can be biased on training data~\cite{loecher_debiasing_2022}. For feature effects, this issue is largely unstudied. \ac{PD}~\cite{friedman_greedy_2001}, \ac{ALE}~\cite{apley_visualizing_2020}, and common references and software often use training data without justification~\cite{greenwell_pdp_2017,molnar_interpretable_2022}, while others use holdout data~\cite{molnar_relating_2023}.
Practitioners likewise disagree: some prefer training data for more reliable estimates due to larger sample size\footnote{\url{https://forums.fast.ai/t/partial-dependence-plot/98465} (last accessed: 01/28/2026)}, others prefer holdout data as ``the model may overfit''\footnote{\url{https://github.com/sosuneko/PDPbox/issues/68} (01/28/2026)}, and some use both to diagnose distribution differences.\footnote{\url{https://www.mathworks.com/help/stats/use-partial-dependence-plots-to-interpret-regression-models-trained-in-regression-learner-app.html} (01/28/2026)}
To date, no systematic study addresses this. We do so by comparing different estimation strategies (training/holdout/\ac{CV}) through bias-variance decomposition.
%
%
%
%%%%%%%%%%%%%%%%%%%%%%%%%%%%%%%%%%%%%%%%%%%%%%%%%%%%%%%%%%%%%%%%%%
%
%                           Background
%
%%%%%%%%%%%%%%%%%%%%%%%%%%%%%%%%%%%%%%%%%%%%%%%%%%%%%%%%%%%%%%%%%%
\section{Notation \& Background}\label{sec:background}

Assume a data-generating process that is characterized by a joint distribution $\Pxy$ over features $\X=(X_1, \dots, X_p)^\top$ and target $\Y$, with true underlying function $\ftrue(x)=\mathbb{E}[\Y\mid \X=x]$.
A random dataset $\DatasetRV=\{(X^{(i)},Y^{(i)})\}_{i=1}^n$ consists of $n$ i.i.d. samples from $\Pxy$.
Concrete realizations $\Dataset=\{(\xvec^{(i)},y^{(i)})\}_{i=1}^n$ are the training, test, and validation sets $\Dtrain$, $\Dtest$, and $\Dval$.
% \footnote{$\mathcal{D}_\text{train}$ is used for model fitting and $\mathcal{D}_\text{test}$ to obtain a final unbiased estimate of the model's generalization performance. $\mathcal{D}_\text{val}$ is an additional holdout set, e.g., for model selection, hyperparameter tuning, or model explanations, without biasing the final performance estimate.}
A learning algorithm induces a fitted model $\fhat$ on $\Dtrain$, viewed either as a fixed function or as a random variable with distribution $\Pmodel$ due to training-sample and algorithmic randomness.
Model performance is measured by the risk $\RiskL(\fhat)=\mathbb{E}_{XY}[L(\Y,\fhat(\X))]$ with a point-wise loss $L$, and can be estimated on an $\nsamples$-sized dataset $\Dataset$ as empirical risk $\RiskEmp(\fhat; \Dataset)=\frac{1}{\nsamples}\sum_{i=1}^{\nsamples} L\big(y^{(i)}, \fhat(\xvec^{(i)})\big).$ For $\Dtrain$, this measures in-sample, for $\Dtest$ or $\Dval$ out-of-sample error.
Expectations and variances w.r.t. $\fhat \sim \Pmodel$ are denoted by subscript $F$, $\DatasetRV\sim\Pxy$ by $D$, and $(\X,\Y) \sim \Pxy$ by $XY$ (analogously for marginal and conditionals). Note that $\DatasetRV$ and $\fhat$ are independent (denoted $\DatasetRV \perp \fhat$) only if $\DatasetRV$ is not used for training, i.e., only for holdout data.

We consider feature effects for a subset of features $\XS$ with $\Sidx \subseteq \Sset$.
Throughout this work, we restrict attention to a single feature of interest ($|\Sidx|=1$) and denote the complement feature set by $\XnotS$.

\begin{definition}[\ac{PD}~\cite{friedman_greedy_2001}]
    Let $\func : \Xspace \to \Yspace$ be a prediction function and let $\XS$ denote a feature of interest.
    The \ac{PD} of $\func$ w.r.t. $\XS$ is defined as
    \begin{equation}
        \PDsTheo{\func}(\xS)
        :=
        \E_{\XnotS}\!\left[\func(\xS, \XnotS)\right]
        =
        \int_{\Xspace_{\notSidx}} \func(\xS, \xnotS) \,
        d\mathbb{P}_{\XnotS}(\xnotS).
    \end{equation}
    Given an $\nsamples$-sized dataset $\Dataset$, it can be estimated via Monte Carlo integration as
    \smash{
        $\PDsEst{\func}(\xS)
        :=
        \frac{1}{\nsamples} \sum_{i=1}^{\nsamples} \func(\xS, \xnotS^{(i)})
        = 
        \frac{1}{\nsamples} \sum_{i=1}^{\nsamples} \func^{(i)}_S(x_S)$.
    }
\end{definition}

Here, \smash{$\func^{(i)}_S(x_S)=\func(\xS,\xnotS^{(i)})$} are the \ac{ICE} curves~\cite{goldstein_peeking_2015}. 
$\PDsTheo{\fhat}$ denotes the theoretical \ac{PD} of the model $\fhat$, and $\PDsEst{\fhat}$ is its estimator. $\PDsTheo{\ftrue}$ is the ground-truth \ac{PD}. In practice, \ac{PD} is visualized using a grid of $G$ feature values \smash{$\{\xS^{(g)}\}_{g=1}^G$}. Quantile-based grids rather than equidistant ones are recommended in~\cite{molnar_general_2022}. 

\begin{definition}[\ac{ALE}~\cite{apley_visualizing_2020}]
    Let $\func : \Xspace \to \Yspace$ be a prediction function and let $\XS$ denote a feature of interest.
    The uncentered \ac{ALE} of $\func$ w.r.t. $\XS$ is defined as
    \begin{equation}\label{eq:ale-theoretical-uncentered}
        \uALEsTheo{\func}(\xS)
        =
        \lim_{\Kbins\to\infty}
        \sum_{\kbin=1}^{\ksKx}
        \E_{\XnotS\,|\,\XS \in \IsKk}
        \left[\FinDiffSK{\func}(\kbin, \XnotS)\right],
    \end{equation}
    where feature $\XS$ is partitioned into $\Kbins$ intervals
    $\{\IsKk=(\zsKlower, \zsKupper]\}_{\kbin=1}^{\Kbins}$,
    $\ksKx$
    denotes the index of the interval into which a value $\xS$ falls,
    and the maximum interval width converges to zero as $\Kbins \to \infty$.
    The finite differences in the $\kbin$-th interval are given by
    $
    \FinDiffSK{\func}(\kbin, \xnotS)
    :=
    \func(\zsKupper, \xnotS)
    -
    \func(\zsKlower, \xnotS)
    $.

    The uncentered \ac{ALE} for a finite dataset $\Dataset$ and finite $\Kbins$ can be estimated as follows, where $\nsKk$ denotes the number of observations in the $\kbin$-th interval:
        $\uALEsEst{\func}(\xS) =
        \sum_{\kbin=1}^{\ksKx} \frac{1}{\nsKk} \sum_{i:\xS^{(i)} \in \IsKk}
        \left[\FinDiffSK{\func}(\kbin, \xnotS^{(i)})\right]$.
\end{definition}

Centered versions $\cALEsTheo{h}$ and $\cALEsEst{h}$ are obtained by subtracting a constant such that they have zero mean w.r.t. the marginal distribution of $\XS$ or the empirical distribution of $\{\xS^{(i)}\}_{i=1}^{\nsamples}$, respectively.
For the grid $\{\zsKupper\}_{\kbin=0}^{\Kbins}$, empirical quantiles of $\{\xS^{(i)}\}_{i=1}^{\nsamples}$ are recommended~\cite{apley_visualizing_2020}.
For readability, we omit the superscript $\Kbins$ when $\Kbins$ is fixed, and suppress the subscript $S$ for \ac{PD} and \ac{ALE}.

In~\cite{molnar_relating_2023}, it is shown that the \ac{MSE} of the theoretical \ac{PD} of $\fhat$ w.r.t. to the theoretical ground-truth  \ac{PD} can be decomposed into squared bias and variance:
\begin{equation}\label{eq:pd-decomposition}
    \E_{F}[{(\PDTheo{\ftrue}(\xS) - \PDTheo{\fhat}(\xS))}^2]
    = {(\PDTheo{\ftrue}(\xS) - \E_{F}[\PDTheo{\fhat}(\xS)])}^2 + \text{Var}_{F}[\PDTheo{\fhat}(\xS)].
\end{equation}
The bias term relates to systematic model bias, and the variance term captures variability across model fits.
For the empirical estimator $\PDEst{\fhat}$, Molnar et al.~\cite{molnar_relating_2023} further argue that Monte Carlo integration introduces an additional source of variance.
Moreover, when estimated on holdout data, $\PDEst{\fhat}$ is unbiased w.r.t. $\PDTheo{\fhat}$, and unbiasedness of the model implies unbiasedness of the \ac{PD}.\footnote{Proofs can be found in Appendices C \& D of the arXiv version of \cite{molnar_relating_2023}.}

While an analogous decomposition for \ac{ALE} is not available, consistency results for the \ac{ALE} estimator exist in~\cite{apley_visualizing_2020}. To this end, they define a population version of the binned uncentered \ac{ALE} for a prediction function $\func$ as
    $\uALETheoBinned{\func}(\xS)
    :=
    \sum_{\kbin=1}^{\ksx}
    \E_{\XnotS \,|\, \XS\in \Isk}\!\left[
    \FinDiff{\func}(\kbin, \XnotS)
    \right]$.
For the same $\Kbins$ fixed bins, $\uALEEst{\func}$ on an i.i.d. $\DatasetRV\sim\Pxy$ converges to $\uALETheoBinned{\func}$ pointwise as $\nsamples\to\infty$ almost surely, under mild integrability conditions.
Moreover, $\uALETheoBinned{\func}$ converges pointwise to $\uALETheo{\func}$ as the bin resolution $\Kbins\to\infty$.
Thus, $\uALEEst{\func}$ is jointly consistent for $\uALETheo{\func}$ if $\nsamples$ grows sufficiently fast relative to $\Kbins$.\footnote{These consistency results can be found in Theorem 3 of the arXiv version of \cite{apley_visualizing_2020}.}
%
%
%
%%%%%%%%%%%%%%%%%%%%%%%%%%%%%%%%%%%%%%%%%%%%%%%%%%%%%%%%%%%%%%%%%%
%
%                             Theory
%
%%%%%%%%%%%%%%%%%%%%%%%%%%%%%%%%%%%%%%%%%%%%%%%%%%%%%%%%%%%%%%%%%%
\section{Theoretical Considerations \& Estimators}\label{sec:theoretical-framework}

For our theoretical analysis, we adopt several assumptions listed in \S\ref{app:assumptions}.

\subsection{Full Error Decomposition of the \ac{PD} Estimator}\label{sec:theory-pd}
Previous work~\cite{molnar_relating_2023} considers only the error decomposition of $\PDTheo{\fhat}$. Since $\PDTheo{\fhat}$ cannot be determined when $\mathbb{P}_{\XnotS}$ is unknown and is estimated via Monte Carlo integration, we instead study the estimator's error. While~\cite{molnar_relating_2023} notes that this adds a variance term, we formally derive the \ac{MSE} decomposition at fixed $\xS$, integrating over both $\hat f\sim\Pmodel$ and the dataset $D\sim\Pxy$ used for \ac{PD} estimation: 
\begin{align}
    \begin{split}
        \E_F\E_{\DatasetRV\mid\fhat}\big[{(\PDTheo{\ftrue}(\xS) - \PDEst{\fhat}(\xS))}^2\big]
       = {(\PDTheo{\ftrue}(\xS) - \E_F\E_{\DatasetRV\mid\fhat}[\PDEst{\fhat}(\xS)])}^2 \\
       + \text{Var}_F\E_{\DatasetRV\mid\fhat}[\PDEst{\fhat}(\xS)] 
        + \E_F\text{Var}_{\DatasetRV\mid\fhat}[\PDEst{\fhat}(\xS)].
    \end{split}
    \label{eq:full-pd-decomposition}
\end{align}
The proof is given in \S\ref{app:proof-decomposition}. The decomposition has three distinct terms, which we will analyze in more detail below: the squared bias and two variances.

\textbf{Bias.} The bias of the \ac{PD} estimator (cf. first term in Eq.~\eqref{eq:full-pd-decomposition}) decomposes as:
\begin{align}
    \begin{split}
        \E_{F}\E_{\DatasetRV\mid\fhat}[\PDEst{\fhat}(\xS)] - \PDTheo{\ftrue}(\xS)
        = \\
        \E_{F}[\E_{\DatasetRV\mid\fhat}[\PDEst{\fhat}(\xS)] - \PDTheo{\fhat}(\xS)] + (\E_{F}[\PDTheo{\fhat}(\xS)] - \PDTheo{\ftrue}(\xS))
    \end{split}
\end{align}
by adding and subtracting $\E_{F}[\PDsTheo{\fhat}(\xS)]$.
The first part vanishes by the unbiasedness of the \ac{PD} estimator w.r.t. the theoretical model \ac{PD}~\cite{molnar_relating_2023}, if only the data used for Monte Carlo integration in the \ac{PD} estimation is independent of the model \smash{$\fhat$}. 
This is true for holdout data, but not necessarily for training data.
For the second part, exchanging expectations (Fubini) yields (proof in~\S\ref{app:proof-pd-bias}):
\begin{equation}\label{eq:pd-model-bias}
\E_{F}\!\big[\PDTheo{\fhat}(\xS)\big] - \PDTheo{\ftrue}(\xS)
=
\E_{\XnotS}\!\left[
\E_{F}\!\big[\fhat(\xS,\XnotS)\big] - \ftrue(\xS,\XnotS)
\right].
\end{equation}
Consequently, the term reduces to the model's bias averaged over $\mathbb P_{\XnotS}$.
Thus, for estimation on holdout data, the \ac{PD} estimator's bias reduces to the average model bias, whereas estimation on training data may introduce additional bias.

\textbf{Model variance.}
The second term in Eq.~\eqref{eq:full-pd-decomposition} reflects the variance of the \ac{PD} estimator w.r.t. the model distribution $\Pmodel$.
When the \ac{PD} is estimated on holdout data, it reduces to 
$
\Var_{F}[\E_{\DatasetRV\mid \fhat}[\PDEst{\fhat}(\xS)]] = \Var_{F}[\PDTheo{\fhat}(\xS)]
$
by the unbiasedness of the \ac{PD} estimator w.r.t. the theoretical model \ac{PD}.
This is exactly the variance of the theoretical \ac{PD} in Eq.~\eqref{eq:pd-decomposition}.
By exchanging expectations (Fubini/Tonelli) and applying Jensen's inequality, we obtain an upper bound (proof in \S\ref{app:proof-pd-theoretical-variance}):
\begin{equation}\label{eq:pd-model-variance}
    \Var_{F}[\PDTheo{\fhat}(\xS)] \leq \E_{\XnotS}\!\Var_{F}[\fhat(\xS, \XnotS)].
\end{equation}
Thus, the theoretical \ac{PD} variance at $\xS$ is controlled by the average (pointwise) model variance w.r.t. the marginal distribution of $\XnotS$.

\begin{figure}[b]
    \centering
    \includegraphics[width=1\linewidth]{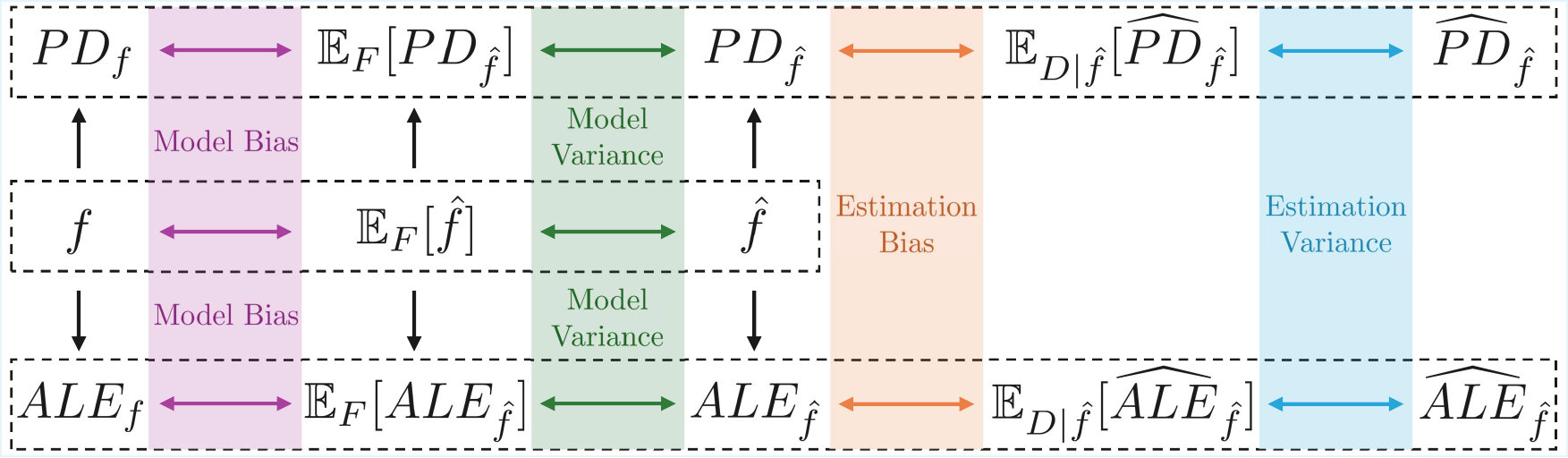}
    \caption{Conceptual overview of the four error components.}
    \label{fig:error-overview}
\end{figure}

\textbf{Estimation variance.}
The third term in Eq.~\eqref{eq:full-pd-decomposition} captures the variance w.r.t. the samples used to estimate the \ac{PD} via Monte Carlo integration, which equals:
\begin{equation}\label{eq:pd-estimation-variance}
    \E_{F}\Var_{\DatasetRV\mid \fhat}[\PDEst{\fhat}(\xS)]
    =\frac{1}{\nsamples}\E_{F}\Var_{\DatasetRV\mid\fhat}[\fhat(\xS,\XnotS)]
\end{equation}
for any $\XnotS \sim \DatasetRV|\fhat$.
The proof is given in~\S\ref{app:proof-pd-mc-variance}.
Consequently, the pointwise estimation variance of the \ac{PD} depends on the sample size $n$ and decreases at the rate $\mathcal O(1/n)$.
It also depends on the expected variance of the \ac{ICE} curves at $\xS$.
Centering the \ac{PD} via $\PDEst{\fhat}(\xS) - \E_{\DatasetRV\mid\fhat}[\PDEst{\fhat}(\XS)]$ yields the variance of the centered \ac{ICE} curves $\Var_{\DatasetRV\mid\fhat} [\fhat(\xS, \XnotS)-\E_{\DatasetRV\mid\fhat}[\fhat(\XS, \XnotS)]]$ instead (proof analogous to \S\ref{app:proof-pd-mc-variance}).
% Note: Now in the next step, we use local decomposability for conditional variance on \fhat, although it was defined/proven in \cite{herbinger_decomposing_2024} only for fixed \fhat, but I think this is the same / this constitutes no problem.
Since centered \ac{ICE} curves fulfill local decomposability~\cite{herbinger_decomposing_2024}, their variance at $\xS$ is solely due to interactions involving feature $\XS$.
Thus, the estimation variance of the centered \ac{PD} is zero when $\XS$ has no interactions in $\fhat$. An overview of all four derived error components is provided in Fig.~\ref{fig:error-overview}.

\subsection{Full Error Decomposition of the \ac{ALE} Estimator}\label{sec:theory-ale}

% As stated in~\cite{molnar_relating_2023}, the results for \ac{PD} also apply to its conditional variants.
% In contrast, \ac{ALE} is not simply a conditional \ac{PD}, as it is defined via accumulated local finite differences, so we provide a dedicated error analysis for \ac{ALE}.
We now provide an error analysis for \ac{ALE}, which is missing in the current literature.
The \ac{MSE} of $\cALEEst{\fhat}$ w.r.t. $\cALETheo{\ftrue}$ can be decomposed into bias and variance analogous to \ac{PD} in
Eq.~\eqref{eq:full-pd-decomposition} (proof in \S\ref{app:proof-decomposition}, analogous to the one for \ac{PD}).
The only property required for the argument is that all expectations exist and that $\ftrue$, and hence $\cALETheo{\ftrue}$, is non-random w.r.t. to $\Pmodel$.
For the theoretical analysis of the error components, we focus on uncentered \ac{ALE}, as centering~is linear post-processing that only affects the offset. All sources of bias and variance originate in the uncentered \ac{ALE} and propagate deterministically under centering.

\textbf{Bias.} Similar to \ac{PD}, we can decompose the bias of \ac{ALE} further into:
\begin{equation}\label{eq:ale-bias-decomp}
    \begin{split}
        \E_{F}\E_{\DatasetRV\mid \fhat}[\uALEEst{\fhat}(\xS)]
        - \uALETheo{\ftrue}(\xS)
        =\\
        \E_{F}\!\left[
        \E_{\DatasetRV\mid \fhat}[\uALEEst{\fhat}(\xS)]
        - \uALETheo{\fhat}(\xS)
        \right]
        + \big(
        \E_F[\uALETheo{\fhat}(\xS)]
        - \uALETheo{\ftrue}(\xS)\big),
    \end{split}
\end{equation}
by adding and subtracting $\E_{F}[\uALETheo{\fhat}(\xS)]$.
The first part can be viewed as the average ``estimation bias'' and can be further decomposed into:
\begin{equation}
\begin{split}
    \E_{F}[\E_{\DatasetRV\mid \fhat}[\uALEEst{\fhat}(\xS)] - \uALETheo{\fhat}(\xS)] =\\
    \E_{F}[\E_{\DatasetRV\mid \fhat}[\uALEEst{\fhat}(\xS)] - \uALETheoBinned{\fhat}(\xS)] + \E_{F}[\uALETheoBinned{\fhat}(\xS) - \uALETheo{\fhat}(\xS)].
\end{split}
\end{equation}
The first term is due to estimation on finite samples.
For \ac{ALE} estimation on holdout data, under mild regularity (Assumptions~\ref{ass:fin-cond-exp} and \ref{ass:delta-moments}) and $\nsk>0$ for all relevant bins, the \ac{ALE} estimator is unbiased w.r.t. the binned population \ac{ALE} and the term becomes zero (proof in \S\ref{app:proof-ale-estimation-bias}).
To the second term, we refer as ``discretization bias'': the term inside $\E_{F}$ goes to 0 as $\Kbins\to\infty$ (pointwise in $\fhat$) by the definition of \ac{ALE}~(Eq.~\eqref{eq:ale-theoretical-uncentered}).
Convergence of the entire expectation term can be shown by applying the dominated convergence theorem exactly as in \S\ref{app:proof-ale-model-bias}.

For the second part in Eq.~\eqref{eq:ale-bias-decomp}, we show in \S\ref{app:proof-ale-model-bias} that it arises from an (infinite) sum of the local average biases in the finite differences of $\fhat$ w.r.t. those of $\ftrue$.
Thus, it vanishes when the model's conditional finite differences are unbiased w.r.t. those of $\ftrue$. In particular, this is satisfied if $\fhat$ is unbiased w.r.t. $\ftrue$ for all $\xvec$.

\textbf{Model variance.}
When \ac{ALE} is estimated on holdout data, it holds that $\Var_{F}[\E_{\DatasetRV\mid\fhat}[\uALEEst{\fhat}(\xS)]] = \Var_{F}[\uALETheoBinned{\fhat}(\xS)]$ by the \ac{ALE} estimator's unbiasedness (see above).
For this expression, we obtain an upper bound for the variance of \ac{ALE} w.r.t. the model distribution (proof in \S\ref{app:proof-ale-model-variance}):
\begin{equation}\label{eq:ale-model-variance-bound}
    \Var_{F}\!\left[\uALETheoBinned{\fhat}(\xS)\right]
    \leq
    \ksx \!\! \sum_{\kbin=1}^{\ksx}
    \E_{\XnotS\mid \XS \in \Isk}
    \Var_{F}
    \!\left[
    \FinDiff{\fhat}(\kbin, \XnotS)
    \right].
\end{equation}
Thus, the theoretical \ac{ALE} variance at $\xS$ is controlled by the local average variability of finite differences across models along the intervals up to $\xS$. 

\textbf{Estimation variance.}
By conditioning on the samples' bin assignments, $\BBinIndex=(\BBinIndex^{(1)},\ldots,\BBinIndex^{(n)})$ with $\BBinIndex^{(i)}=k_{\Sidx}(\XS^{(i)})$, and using the law of total variance, the estimation variance of \ac{ALE}, $\E_{F}[\Var_{\DatasetRV\mid\fhat}[\uALEEst{\fhat}(\xS)]]$, decomposes into
\begin{align*}
    \E_{F}\!\left[
    \Var_{\BBinIndex\mid\fhat}\!\left[
    \E_{\DatasetRV\mid\fhat,\BBinIndex}\!\left[\uALEEst{\fhat}(\xS)\right]
    \right]
    \right]
    + 
     \E_{F}\!\left[
    \E_{\BBinIndex\mid\fhat}\!\left[
    \Var_{\DatasetRV\mid\fhat,\BBinIndex}\!\left[\uALEEst{\fhat}(\xS)\right]
    \right]
    \right].
\end{align*}
The first term captures variance from random bin assignments along $\XS$, i.e., variability in the expected \ac{ALE} estimate due to random sample allocation to bins.
For the second term, assuming $\nsk>0$ for all relevant bins $k\le \ksx$: % (so $\uALEEst{\fhat}(\xS)$ is well-defined):
\begin{equation}\label{eq:ale-estimation-variance-cond}
    \E_{F}\E_{\BBinIndex\mid\fhat}\Var_{\DatasetRV\mid\fhat,\BBinIndex}\!\bigg[\uALEEst{\fhat}(\xS)\bigg]
    =
    \sum_{k\le \ksx} \E_{F}\E_{\BBinIndex\mid\fhat}\left[\frac{1}{\nsk}\sigma_k^2(\fhat)\right]
\end{equation}
with $\sigma_k^2(\fhat):=\Var_{\XnotS\mid \XS\in \Isk, \fhat}[\FinDiff{\fhat}(k,\XnotS)]$. The proof is given in \S\ref{app:proof-ale-estimation-variance}.
Thus, it scales with the expected inverse number of observations per bin.
Assuming deterministic equal-frequency binning for simplicity, this factor becomes $\frac{\Kbins}{\nsamples}$.
It also depends on the local variance in the estimated finite differences.
% Note: For the following, exactly the same modification of \cite{herbinger_decomposing_2024} is necessary as above after equation~\ref{eq:pd-estimation-variance}, again this is straightforward.
Finite differences also fulfill local decomposability~\cite{herbinger_decomposing_2024}.
Thus, the variance $\sigma_k^2(\fhat)$ is only due to interaction effects involving feature $\XS$ and is zero when $\XS$ has no interaction effects in $\fhat$.

\subsection{Estimators of the Error Components}

To investigate the error components empirically in \S\ref{sec:set-up}-\ref{sec:results}, we propose estimators for each.
$\E_F$ and $\Var_F$ can be estimated by averaging over multiple models $\fhat^{(m)}$ of the same learner fitted to $M$ different training sets, each independently sampled from $\Pxy$.
Translating the standard estimators for \ac{MSE}, bias, and variance~\cite{morris_using_2019} to our setting yields the estimators in Tab.~\ref{tab:error-component-estimators}.
The variance estimator is the same as in~\cite{molnar_relating_2023}, capturing the variance in both model fits and Monte Carlo integration.
To estimate the estimation variance separately, consider $R$ Monte Carlo iterations $\mathcal{D}_r$ to estimate the feature effect $\PDEst{}^{[r]}$, again for multiple model refits.
While Tab.~\ref{tab:error-component-estimators} is based on \ac{PD}, estimators for \ac{ALE} follow analogously.

\begin{table}[tb]
    \centering
    \scriptsize
    \caption{Estimators of the feature effect error components at $\xS$.}
    \label{tab:error-component-estimators}
    \begin{tabularx}{\textwidth}{@{}l X@{}}
        \toprule
        \textbf{Component} & \textbf{Estimator} \\
        \midrule
        \ac{MSE} & $\widehat{\text{MSE}}(\xS) = \frac{1}{M} \sum_{m=1}^{M} {\big(\PDTheo{\ftrue}(\xS) - \PDEst{\fhat^{(m)}}(\xS)\big)}^2$ \\
        Bias & $\widehat{\text{Bias}}(\xS) = \PDTheo{\ftrue}(\xS) - \frac{1}{M}\sum_{m=1}^M \PDEst{\fhat^{(m)}}(\xS)$ \\
        Variance & $\widehat{\text{Var}}(\xS) = \frac{1}{M-1}\sum_{m=1}^M{\left(\PDEst{\fhat^{(m)}}(\xS) - \frac{1}{M}\sum_{m'=1}^M \PDEst{\fhat^{(m')}}(\xS)\right)}^2$ \\
        Estimation variance & $\widehat{\Var}_\text{Est}(\xS) = \frac{1}{M(R-1)}\sum_{m=1}^M \sum_{r=1}^R {\big(\PDEst{\fhat^{(m)}}^{[r]}(\xS) - \frac{1}{R}\sum_{r'=1}^R \PDEst{\fhat^{(m)}}^{[r']}(\xS)\big)}^2$ \\
        \bottomrule
    \end{tabularx}
\end{table}

These estimators are unbiased according to standard statistical results. %~\cite{casella_statistical_2024}.
The Monte Carlo standard errors of these estimates can again be estimated as a function of the number of repetitions~\cite{morris_using_2019}.
Importantly, most estimators in Tab.~\ref{tab:error-component-estimators} are impractical as they require knowledge of the ground-truth.
We will rather use them to understand the behavior of feature effect estimation.
%
%
%
%%%%%%%%%%%%%%%%%%%%%%%%%%%%%%%%%%%%%%%%%%%%%%%%%%%%%%%%%%%%%%%%%%
%
%                           Experiments
%
%%%%%%%%%%%%%%%%%%%%%%%%%%%%%%%%%%%%%%%%%%%%%%%%%%%%%%%%%%%%%%%%%%
\section{Experimental Set-Up}\label{sec:set-up}

We now empirically validate our findings from \S\ref{sec:theoretical-framework} with an extensive simulation study. While the theoretical analysis requires holdout data at multiple points (e.g., to avoid estimation bias), we aim to gain deeper insights into what empirically happens when this is violated. For this, we compare feature effect errors across different estimation strategies: on training data, validation data, and via \ac{CV}. We break this overarching goal down into three specific research questions:
\begin{description}[topsep=5pt]
    \item[RQ1:] How does overfitting empirically affect \ac{MSE}, bias, and variance of \ac{PD} and \ac{ALE} when estimated on training vs. validation data vs. \ac{CV}?\label{rq:1}
    \item[RQ2:] How do the variance sources (model and estimation variance) empirically behave for \ac{PD} and \ac{ALE} on training vs. validation data vs. in \ac{CV}?\label{rq:2}
    \item[RQ3:] How does sample size affect estimation error for \ac{PD} and \ac{ALE} empirically?\label{rq:3}
\end{description}

\textbf{Data settings.}
We consider three settings of varying complexity. 
Ground-truth functions and feature structures are given in Tab.~\ref{tab:data-settings}. 
All settings include two independent dummy features. 
The first setting has correlations and interactions, the second different non-linearities, and the third real-world relevance.\footnote{This dataset is based on the physics-grounded Feynman equation I.29.16 for wave interference, addressing the limited realism of standard test functions.}
Features are generated i.i.d., and the target as $y=\ftrue(\mathbf{x})+\varepsilon$ with i.i.d.\ $\varepsilon \sim\mathcal{N}(0,\sigma_\varepsilon^2)$ and a signal-to-noise ratio of $5$.\footnote{The scale parameter is set such that $\hat\sigma_Y/\sigma_\varepsilon=5$.}
% To determine the noise scaling factor, we compute the standard deviation of the signal over 100,000 randomly drawn samples of $y$ and divide by the signal-to-noise ratio of five.
We consider two sample sizes $n=1250$ and $n=10000$.

\begin{table}[tb] % tb
    \centering
    \scriptsize
    \caption{Data settings with underlying functions and feature structure.}
    \begin{tabular}{p{0.19\textwidth} p{0.43\textwidth} p{0.34\textwidth}}
        \toprule
        \textbf{Setting} & \textbf{Function} & \textbf{Correlation}\\
        \midrule
        \textbf{Simple-Normal-Correlated} &
        \(
        \begin{array}{l}
        f_1(\mathbf{x}) = x_1 + \frac{x_2^2}{2} + x_1 x_2
        \end{array}
        \) &
        \(
        \begin{array}{l}
        X_1,\dots,X_4\sim N\left(0,1\right),  \rho_{12}=0.9, \\ \rho_{ij}=0 \; \forall (i,j) \neq (1,2),\ i \neq j
        \end{array}
        \)
        \\
        \midrule
        \textbf{Friedman1} \cite{friedman_multivariate_1991} &
        \(
        \begin{array}{l}
        f_2(\mathbf{x}) = 10 \sin(\pi x_1 x_2) \\+ 20{(x_3 - \frac{1}{2})}^2 + 10 x_4 + 5 x_5
        \end{array}
        \) & $X_1, \dots, X_7 \overset{\text{i.i.d.}}{\sim} \mathrm{U}(0,1)$ \\
        \midrule
        \textbf{Feynman I.29.16} \cite{matsubara_rethinking_2024} &
        \(
        \begin{array}{l}
        f_3(\mathbf{x}, \boldsymbol{\theta}) = \sqrt{x_1^2 + x_2^2 + 2 x_1 x_2 \cos(\theta_1 - \theta_2)}
        \end{array}
        \) &
        \(
        \begin{array}{l}
        X_1,X_2\sim \mathrm{LogU}(0.1,10), \\
        \theta_1,\theta_2\sim \mathrm{U}(0, 2\pi), \\
        D_1,D_2\sim \mathrm{U}(0, 1), \text{ all indep.}
        \end{array}
        \)\\
         \bottomrule
    \end{tabular}
    \label{tab:data-settings}
\end{table}

\textbf{Models.}
We consider a \ac{GAM} with spline bases for main and pairwise interaction effects,
and XGBoost as learners.
Each of them is once configured with ``optimally tuned'' (OT) hyperparameters and once with hyperparameters chosen to overfit (OF).
These hyperparameters are pre-selected per setting and sample size, i.e., carefully hand-picked for OF (e.g., small penalty / large learning rate) and tuned on separate data samples for OT.
For details on the hyperparameters and model performances, see \S\ref{app:model-hps}.%  and \S\ref{app:model-performance}.

\textbf{Feature effect estimation.}
We estimate the feature effects $\PDEst{\fhat}$ and $\cALEEst{\fhat}$ per feature and model by (a) training a model on all $n$ samples and estimating the feature effect on the same $n$ samples, (b) splitting the $n$ samples into $80\%$ train and $20\%$ validation set, fitting a model on the training set and estimating the effect on the validation set,  and (c) a \ac{CV}-based estimation strategy on all $n$ samples.\footnote{We use 5-fold \ac{CV}: in each iteration, a model is fitted on four folds, effects are estimated on the held-out fold, and the five resulting effects are averaged pointwise.}
We compute feature effects all at the same 100 grid points, defined by the theoretical quantiles of $\mathbb{P}_{X_\Sidx}$ for comparability. 
Additionally, we center the curves after estimation,
%to have a mean effect of $0$ w.r.t.\ the grid points,
and omit the first and last grid point to avoid boundary effects, particularly for \ac{ALE}.
To enable ground-truth comparisons, we additionally construct ground-truth effect estimators in the same manner, but 
on $10,000$ fresh samples and directly on $\ftrue$.
%, as the complexity of \textit{Feynman I.29.16} makes an analytical computation infeasible. 
This eliminates the discretization bias from our error, as the ground-truth and estimate use the same intervals.\footnote{Except for discretization bias, this estimate is unbiased w.r.t. the true theoretical effect (cf.~\S\ref{sec:theoretical-framework}), and adds negligible estimation variance at this sample size (cf.~\S\ref{sec:results-sample-size}).}

\textbf{Experiments.} To address RQ1, we compute the \ac{MSE}, bias, and variance of all estimated model feature effects w.r.t. the estimated ground-truth effects via the estimators in Tab.~\ref{tab:error-component-estimators} with $\PDEst{\ftrue}$ for $\PDTheo{\ftrue}$ (analog. for ALE).
We repeat each setting-size-model combination $M=30$ times to estimate the error terms. Each repetition involves drawing new data, fitting the models, and estimating the effects. 
We report \ac{MSE}, bias, and variance averaged over the grid points.

To address RQ2, we estimate the estimation variance according to Tab.~\ref{tab:error-component-estimators}. 
In each iteration $m\in\{1,\ldots,M\}$, we fix the trained models, draw $R=30$ new data sets, and estimate the effects with them for the fixed models.
By subtracting this from the total variance (cf. RQ1), we estimate the model variance. For this analysis, we focus on XGBoost (OF \& OT) and a sample size of $n=1250$.

To address RQ3, we compute the \ac{MSE} between analytical ground-truth effects (available for the first two settings) and estimated ground-truth effects across 50 different sample sizes ranging from $10^1$ to $10^6$ (on log scale) with 50 repetitions per size. %, each on fresh samples.
This isolates the estimation error, as no model is involved.
%We restrict this analysis to the first two data settings, as computing the \textit{FeynmanI.29.16} ground-truth feature effects analytically is infeasible.

\textbf{Reproducibility.}
Experiments are implemented in Python and use fixed random seeds.
We release all experimental code and raw results on GitHub (\href{https://github.com/slds-lmu/paper_2026_xai_feature_effects_code}{link}). 
%
%
%
%%%%%%%%%%%%%%%%%%%%%%%%%%%%%%%%%%%%%%%%%%%%%%%%%%%%%%%%%%%%%%%%%%
%
%                             Results
%
%%%%%%%%%%%%%%%%%%%%%%%%%%%%%%%%%%%%%%%%%%%%%%%%%%%%%%%%%%%%%%%%%%
\section{Empirical Results}\label{sec:results}

We report results for a single setting and a representative feature subset. 
Further results are provided in \S\ref{app:results-bias-variance} and consistent with the findings reported here.

\subsection{Bias-Variance-Analysis}\label{sec:results-bias-variance}

\textbf{\ac{PD} decomposition.}
Our empirical results on \textit{SimpleNormalCorrelated} in Tab.~\ref{tab:res-pd-snc} reveal that the \ac{MSE} of the \ac{PD} estimator is lowest mostly for \ac{CV}- and training-set-based estimation.
Bias is mostly similar across the estimation strategies, and we observe no systematic trends w.r.t. estimation strategy or sample size. 
Thus, a potential bias introduced by estimation on training data that could not be ruled out in our theoretical analysis (\S\ref{sec:theory-pd}) appears empirically negligible.
Variance is generally lowest for \ac{CV}-based estimation. 
We hypothesize that this is due to two effects: (1) model variance may decrease as \ac{CV} averages out fitting variability across multiple models, and (2) estimation variance may decrease compared to estimation on a single validation set due to increased effective sample size.
Variance is generally slightly higher for validation than for training-set-based estimation, likely due to the smaller sample size. 
The higher variance of overfitting models reflects in higher \ac{PD} variance.
These empirical results agree with our theoretical findings. 
An additional finding is that, for well-generalizing models, differences in \ac{MSE} across estimation strategies are mostly negligible, while for overfitting models, \ac{CV}-based estimation yields a substantial reduction.
\begin{table}[t]
    \centering
    \scriptsize
    \caption{Results for \ac{PD} on \textit{SimpleNormalCorrelated} averaged over 100 grid points. Bold numbers are minimum per metric-feature-model-size-combination.}
   \begin{tabular}{lll|rrr|rrr|rrr}
   \toprule
     &  & \textbf{Feature} & \multicolumn{3}{c}{$\mathbf{x_1}$} & \multicolumn{3}{c}{$\mathbf{x_2}$} & \multicolumn{3}{c}{$\mathbf{x_3}$} \\
     &  & \textbf{Metric} & MSE & Bias & Var & MSE & Bias & Var & MSE & Bias & Var \\
    \midrule
\multirow[c]{12}{*}{\rotatebox{90}{$n=1250$}} & \multirow[c]{3}{*}{GAM\_OF} & train & 0.0695 & 0.0296 & 0.0710 & 0.0699 & \bfseries 0.0276 & 0.0716 & 0.0090 & 0.0161 & 0.0091 \\
 &  & val & 0.0928 & \bfseries 0.0206 & 0.0955 & 0.0849 & 0.0345 & 0.0865 & 0.0114 & 0.0190 & 0.0114 \\
 &  & CV & \bfseries 0.0609 & 0.0264 & \bfseries 0.0622 & \bfseries 0.0562 & 0.0292 & \bfseries 0.0573 & \bfseries 0.0086 & \bfseries 0.0157 & \bfseries 0.0086 \\
	\cmidrule(){2-12}
 & \multirow[c]{3}{*}{GAM\_OT} & train & \bfseries 0.0051 & \bfseries 0.0323 & 0.0042 & \bfseries 0.0039 & \bfseries 0.0247 & \bfseries 0.0034 & 0.0005 & \bfseries 0.0022 & 0.0005 \\
 &  & val & 0.0089 & 0.0336 & 0.0081 & 0.0093 & 0.0305 & 0.0086 & 0.0005 & 0.0040 & 0.0005 \\
 &  & CV & 0.0053 & 0.0356 & \bfseries 0.0042 & 0.0041 & 0.0287 & 0.0034 & \bfseries 0.0005 & 0.0022 & \bfseries 0.0005 \\
	\cmidrule(){2-12}
 & \multirow[c]{3}{*}{XGB\_OF} & train & 0.1780 & 0.3147 & 0.0817 & 0.4217 & 0.4505 & 0.2263 & 0.0010 & 0.0043 & 0.0010 \\
 &  & val & 0.1880 & 0.3242 & 0.0857 & 0.3888 & \bfseries 0.3941 & 0.2416 & 0.0015 & 0.0098 & 0.0014 \\
 &  & CV & \bfseries 0.1458 & \bfseries 0.3114 & \bfseries 0.0505 & \bfseries 0.3043 & 0.4236 & \bfseries 0.1291 & \bfseries 0.0007 & \bfseries 0.0042 & \bfseries 0.0007 \\
	\cmidrule(){2-12}
 & \multirow[c]{3}{*}{XGB\_OT} & train & \bfseries 0.2807 & \bfseries 0.5194 & 0.0113 & 0.1690 & 0.3941 & 0.0141 & 0.0014 & 0.0052 & 0.0014 \\
 &  & val & 0.3008 & 0.5375 & 0.0123 & \bfseries 0.1666 & \bfseries 0.3894 & 0.0155 & 0.0019 & 0.0073 & 0.0019 \\
 &  & CV & 0.2950 & 0.5351 & \bfseries 0.0090 & 0.1691 & 0.3987 & \bfseries 0.0104 & \bfseries 0.0013 & \bfseries 0.0051 & \bfseries 0.0013 \\
 \midrule
 \multirow[c]{12}{*}{\rotatebox{90}{$n=10000$}} & \multirow[c]{3}{*}{GAM\_OF} & train & 0.1878 & 0.0674 & 0.1895 & 0.1995 & 0.0890 & 0.1982 & 0.0008 & 0.0050 & 0.0008 \\
 &  & val & 0.2248 & \bfseries 0.0601 & 0.2289 & 0.2371 & \bfseries 0.0678 & 0.2406 & 0.0011 & 0.0055 & 0.0011 \\
 &  & CV & \bfseries 0.1777 & 0.0778 & \bfseries 0.1775 & \bfseries 0.1875 & 0.0996 & \bfseries 0.1837 & \bfseries 0.0008 & \bfseries 0.0050 & \bfseries 0.0008 \\
	\cmidrule(){2-12}
 & \multirow[c]{3}{*}{GAM\_OT} & train & \bfseries 0.0011 & \bfseries 0.0206 & 0.0007 & \bfseries 0.0012 & \bfseries 0.0204 & 0.0008 & 0.0000 & \bfseries 0.0009 & 0.0000 \\
 &  & val & 0.0020 & 0.0228 & 0.0016 & 0.0018 & 0.0222 & 0.0014 & 0.0001 & 0.0012 & 0.0001 \\
 &  & CV & 0.0012 & 0.0232 & \bfseries 0.0007 & 0.0013 & 0.0230 & \bfseries 0.0008 & \bfseries 0.0000 & 0.0010 & \bfseries 0.0000 \\
	\cmidrule(){2-12}
 & \multirow[c]{3}{*}{XGB\_OF} & train & 0.0915 & \bfseries 0.2638 & 0.0227 & 0.2625 & 0.4404 & 0.0710 & 0.0003 & 0.0029 & 0.0003 \\
 &  & val & 0.1029 & 0.2675 & 0.0324 & 0.2700 & 0.4465 & 0.0730 & 0.0004 & 0.0037 & 0.0004 \\
 &  & CV & \bfseries 0.0828 & 0.2641 & \bfseries 0.0135 & \bfseries 0.2226 & \bfseries 0.4277 & \bfseries 0.0411 & \bfseries 0.0002 & \bfseries 0.0022 & \bfseries 0.0002 \\
	\cmidrule(){2-12}
 & \multirow[c]{3}{*}{XGB\_OT} & train & \bfseries 0.1752 & \bfseries 0.4142 & 0.0037 & \bfseries 0.1210 & \bfseries 0.3418 & 0.0043 & 0.0002 & 0.0022 & 0.0002 \\
 &  & val & 0.1802 & 0.4197 & 0.0042 & 0.1226 & 0.3439 & 0.0045 & 0.0003 & 0.0022 & 0.0003 \\
 &  & CV & 0.1806 & 0.4215 & \bfseries 0.0030 & 0.1237 & 0.3471 & \bfseries 0.0033 & \bfseries 0.0002 & \bfseries 0.0021 & \bfseries 0.0002 \\
	\bottomrule
	\end{tabular}
	\label{tab:res-pd-snc}
\end{table}

\textbf{\ac{ALE} decomposition.}
Similar results for \ac{ALE} in Tab.~\ref{tab:res-ale-snc} show that the \ac{MSE} is again lowest for training-set- or \ac{CV}-based estimation.
In contrast to \ac{PD}, bias is often considerably lower for training data (largest set) when $n$ is small, and generally decreases with increasing sample size, likely as the probability of $\nsk>0\ \forall k$ grows (required for unbiasedness, cf. \S\ref{sec:theory-ale}).
Again, the variance is mostly lowest for \ac{CV}-based estimation, but now substantially higher on the smaller validation set, supporting our theoretical result that \ac{ALE} (estimation) variance is more sensitive to sample size than \ac{PD}.
Generally, this confirms our theoretical findings and again shows that \ac{CV}-based estimation is promising.
% However, estimation with the smaller validation set now increases both variance and bias.
%
\begin{table}[t]
    \centering
    \scriptsize
    \caption{Results for \ac{ALE} on \textit{SimpleNormalCorrelated} averaged over 100 grid points. Bold numbers are minimum per metric-feature-model-size-combination.}
   \begin{tabular}{lll|rrr|rrr|rrr}
   \toprule
     &  & \textbf{Feature} & \multicolumn{3}{c}{$\mathbf{x_1}$} & \multicolumn{3}{c}{$\mathbf{x_2}$} & \multicolumn{3}{c}{$\mathbf{x_3}$} \\
     &  & \textbf{Metric} & MSE & Bias & Var & MSE & Bias & Var & MSE & Bias & Var \\
    \midrule
\multirow[c]{12}{*}{\rotatebox{90}{$n=1250$}} & \multirow[c]{3}{*}{GAM\_OF} & train & \bfseries 0.0123 & \bfseries 0.0205 & 0.0123 & \bfseries 0.0108 & \bfseries 0.0236 & \bfseries 0.0106 & 0.0091 & 0.0161 & 0.0092 \\
 &  & val & 0.0277 & 0.0771 & 0.0225 & 0.0304 & 0.0711 & 0.0262 & 0.0129 & 0.0176 & 0.0131 \\
 &  & CV & 0.0178 & 0.0857 & \bfseries 0.0108 & 0.0190 & 0.0858 & 0.0120 & \bfseries 0.0077 & \bfseries 0.0153 & \bfseries 0.0077 \\
	\cmidrule(){2-12}
 & \multirow[c]{3}{*}{GAM\_OT} & train & \bfseries 0.0022 & \bfseries 0.0100 & \bfseries 0.0021 & \bfseries 0.0019 & \bfseries 0.0061 & \bfseries 0.0019 & 0.0005 & 0.0022 & 0.0005 \\
 &  & val & 0.0148 & 0.0808 & 0.0086 & 0.0198 & 0.0918 & 0.0117 & 0.0005 & 0.0036 & 0.0005 \\
 &  & CV & 0.0103 & 0.0899 & 0.0023 & 0.0109 & 0.0890 & 0.0031 & \bfseries 0.0004 & \bfseries 0.0020 & \bfseries 0.0004 \\
	\cmidrule(){2-12}
 & \multirow[c]{3}{*}{XGB\_OF} & train & \bfseries 0.1159 & 0.1343 & \bfseries 0.1012 & 0.1120 & 0.1196 & 0.1011 & 0.0724 & 0.0279 & 0.0741 \\
 &  & val & 0.5238 & 0.1153 & 0.5281 & 0.6017 & 0.1415 & 0.6017 & 0.0848 & 0.0354 & 0.0864 \\
 &  & CV & 0.1531 & \bfseries 0.0784 & 0.1520 & \bfseries 0.1025 & \bfseries 0.0803 & \bfseries 0.0994 & \bfseries 0.0118 & \bfseries 0.0126 & \bfseries 0.0121 \\
	\cmidrule(){2-12}
 & \multirow[c]{3}{*}{XGB\_OT} & train & \bfseries 0.0162 & \bfseries 0.0875 & 0.0088 & \bfseries 0.0174 & \bfseries 0.0768 & 0.0119 & 0.0065 & 0.0077 & 0.0066 \\
 &  & val & 0.0471 & 0.1470 & 0.0263 & 0.0386 & 0.1094 & 0.0275 & 0.0033 & 0.0116 & 0.0033 \\
 &  & CV & 0.0315 & 0.1549 & \bfseries 0.0077 & 0.0234 & 0.1158 & \bfseries 0.0103 & \bfseries 0.0020 & \bfseries 0.0053 & \bfseries 0.0020 \\
 \midrule
\multirow[c]{12}{*}{\rotatebox{90}{$n=10000$}} & \multirow[c]{3}{*}{GAM\_OF} & train & 0.0011 & 0.0071 & 0.0011 & 0.0012 & \bfseries 0.0071 & 0.0011 & 0.0008 & \bfseries 0.0050 & 0.0008 \\
 &  & val & 0.0017 & 0.0075 & 0.0017 & 0.0017 & 0.0085 & 0.0017 & 0.0011 & 0.0057 & 0.0011 \\
 &  & CV & \bfseries 0.0011 & \bfseries 0.0068 & \bfseries 0.0011 & \bfseries 0.0011 & 0.0072 & \bfseries 0.0011 & \bfseries 0.0008 & 0.0050 & \bfseries 0.0008 \\
	\cmidrule(){2-12}
 & \multirow[c]{3}{*}{GAM\_OT} & train & 0.0002 & 0.0030 & 0.0002 & 0.0003 & \bfseries 0.0045 & 0.0002 & \bfseries 0.0000 & \bfseries 0.0009 & 0.0000 \\
 &  & val & 0.0006 & 0.0037 & 0.0006 & 0.0005 & 0.0056 & 0.0005 & 0.0001 & 0.0012 & 0.0001 \\
 &  & CV & \bfseries 0.0002 & \bfseries 0.0029 & \bfseries 0.0002 & \bfseries 0.0003 & 0.0046 & \bfseries 0.0002 & 0.0000 & 0.0010 & \bfseries 0.0000 \\
	\cmidrule(){2-12}
 & \multirow[c]{3}{*}{XGB\_OF} & train & 0.0164 & 0.0346 & 0.0157 & 0.0197 & 0.0520 & 0.0176 & 0.0103 & 0.0189 & 0.0103 \\
 &  & val & 0.0652 & 0.0416 & 0.0657 & 0.0780 & 0.0597 & 0.0770 & 0.0155 & 0.0166 & 0.0157 \\
 &  & CV & \bfseries 0.0130 & \bfseries 0.0180 & \bfseries 0.0131 & \bfseries 0.0147 & \bfseries 0.0198 & \bfseries 0.0148 & \bfseries 0.0023 & \bfseries 0.0098 & \bfseries 0.0023 \\
	\cmidrule(){2-12}
 & \multirow[c]{3}{*}{XGB\_OT} & train & 0.0046 & \bfseries 0.0538 & 0.0017 & 0.0039 & \bfseries 0.0401 & 0.0023 & 0.0005 & 0.0037 & 0.0005 \\
 &  & val & 0.0054 & 0.0544 & 0.0025 & 0.0047 & 0.0412 & 0.0031 & 0.0006 & 0.0037 & 0.0007 \\
 &  & CV & \bfseries 0.0044 & 0.0555 & \bfseries 0.0014 & \bfseries 0.0034 & 0.0417 & \bfseries 0.0017 & \bfseries 0.0003 & \bfseries 0.0027 & \bfseries 0.0003 \\
	\bottomrule
	\end{tabular}
	\label{tab:res-ale-snc}
\end{table}

\subsection{Variance Decomposition Analysis}\label{sec:results-var-decomp}

For variance decomposition into model and estimation variance, we consider the results on \textit{Friedman1} in Tab.~\ref{tab:res-ablation-f1}, including a non-linear feature with interactions ($X_1$), a linear without ($X_4$), and a dummy feature ($X_7$).
The estimation variance is constantly highest when feature effects are estimated on the smaller validation set. 
This is more pronounced for \ac{ALE}, and we generally observe higher estimation variance than for \ac{PD}.
As hypothesized, \ac{CV} (1) reduces model variance compared to the other estimation strategies, which is most pronounced for overfitting models (generally higher model variance). As expected, it also (2) reduces estimation variance compared to a single validation set.
These empirical findings agree with our theoretical results.
Although estimation variance is sometimes slightly higher for features with interactions ($X_1$), XGBoost may also learn interactions that are not in $\ftrue$, which is why this is not as clear as expected.

\begin{table}[tb]
    \centering
    \scriptsize
    \begin{threeparttable}
    \caption{Decomposition results of total variance $\text{Var}_\text{Tot}$ into model variance $\text{Var}_\text{Mod}$ and estimation variance $\text{Var}_\text{Est}$ on \textit{Friedman1} averaged over 100 grid points. Bold numbers indicate estimation strategy with minimal variance.}
    \begin{tabular}{lll|rrr|rrr|rrr}
        \toprule
         &  & \textbf{Feature} & \multicolumn{3}{c}{$\mathbf{x_1}$} & \multicolumn{3}{c}{$\mathbf{x_4}$} & \multicolumn{3}{c}{$\mathbf{x_7}$} \\
         &  & \textbf{Metric} & $\text{Var}_\text{Tot}$ & $\text{Var}_\text{Mod}$ & $\text{Var}_\text{Est}$ & $\text{Var}_\text{Tot}$ & $\text{Var}_\text{Mod}$ & $\text{Var}_\text{Est}$ & $\text{Var}_\text{Tot}$ & $\text{Var}_\text{Mod}$ & $\text{Var}_\text{Est}$ \\
        \midrule
        \multirow[c]{6}{*}{\rotatebox{90}{PD}} & \multirow[c]{3}{*}{XGB\_OF} & train & 0.0862 & 0.0833 & 0.0029 & 0.1269 & 0.1240 & 0.0029 & 0.0027 & 0.0025 & 0.0002 \\
         &  & val & 0.1120 & 0.0974 & 0.0146 & 0.1811 & 0.1661 & 0.0150 & 0.0059 & 0.0046 & 0.0012 \\
         &  & CV & \bfseries 0.0460 & \bfseries 0.0432 & \bfseries 0.0028 & \bfseries 0.0814 & \bfseries 0.0785 & \bfseries 0.0029 & \bfseries 0.0014 & \bfseries 0.0011 & \bfseries 0.0002 \\
          \cmidrule(){2-12}
         &  \multirow[c]{3}{*}{XGB\_OT} & train & 0.0217 & 0.0209 & \bfseries 0.0008 & 0.0232 & 0.0231 & 0.0001 & 0.0071 & 0.0070 & \bfseries 0.0000 \\
         &  & val & 0.0280 & 0.0241 & 0.0038 & 0.0283 & 0.0278 & 0.0006 & 0.0083 & 0.0082 & 0.0001 \\
         &  & CV & \bfseries 0.0190 & \bfseries 0.0182 & 0.0008 & \bfseries 0.0205 & \bfseries 0.0204 & \bfseries 0.0001 & \bfseries 0.0062 & \bfseries 0.0062 & 0.0000 \\
          \midrule
        \multirow[c]{6}{*}{\rotatebox{90}{ALE}} & \multirow[c]{3}{*}{XGB\_OF} & train & 1.0248 & 0.4286 & 0.5962 & 1.1178 & 0.4220 & \bfseries 0.6958 & 0.1852 & 0.1556 & 0.0296 \\
         &  & val & 4.6600 & 1.5201 & 3.1399 & 4.9819 & 1.1399 & 3.8420 & 0.3004 & 0.1614 & 0.1390 \\
         &  &  CV & \bfseries 0.6573 & \bfseries 0.0846 & \bfseries 0.5727 & \bfseries 0.8568 & \bfseries 0.0858 & 0.7710 & \bfseries 0.0262 & \multicolumn{1}{c}{-\tnote{*}} & \bfseries 0.0287 \\
          \cmidrule(){2-12}
         &  \multirow[c]{3}{*}{XGB\_OT} & train & 0.0337 & 0.0229 & \bfseries 0.0108 & 0.0340 & 0.0265 & \bfseries 0.0075 & 0.0125 & 0.0113 & \bfseries 0.0012 \\
         &  & val & 0.0937 & 0.0200 & 0.0737 & 0.1113 & 0.0463 & 0.0651 & 0.0154 & 0.0075 & 0.0080 \\
         &  & CV & \bfseries 0.0299 & \bfseries 0.0147 & 0.0151 & \bfseries 0.0273 & \bfseries 0.0143 & 0.0129 & \bfseries 0.0067 & \bfseries 0.0051 & 0.0016 \\
    	\bottomrule
    \end{tabular}
	\label{tab:res-ablation-f1}
	\begin{tablenotes}
        \item[*] 
        \scriptsize Due to instabilities in variance estimation for ALE, likely caused by unstable bin assignments, the estimated $\text{Var}_\text{Est}$ sometimes exceeds the estimated $\text{Var}_\text{Tot}$. We omit $\text{Var}_\text{Mod}$ in these cases.
    \end{tablenotes}
    \end{threeparttable}
\end{table}

\subsection{Effect of the Sample Size}\label{sec:results-sample-size}

\begin{figure*}[tb]
    \centering
    \begin{subfigure}[b]{0.49\textwidth}
        \centering
        \includegraphics[width=\textwidth]{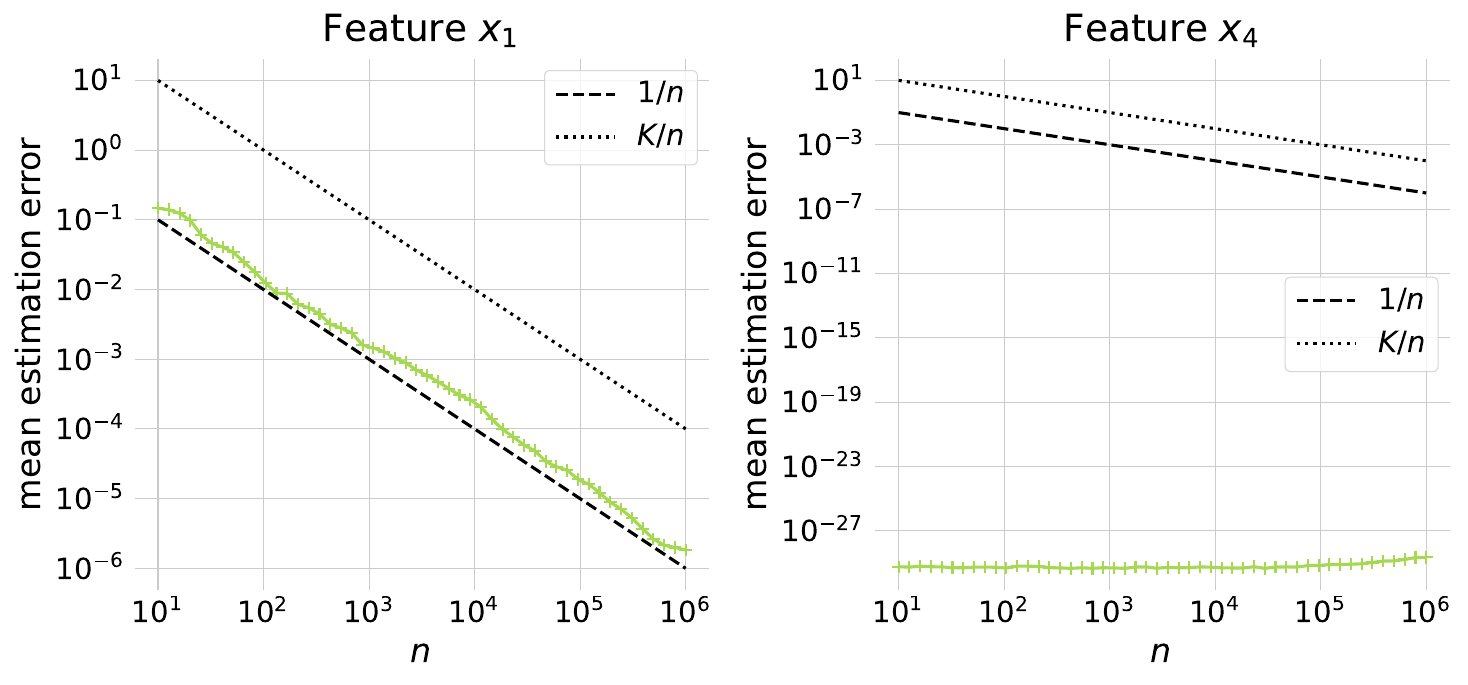}
        \caption{Mean estimation error for \ac{PD}}  
        \label{fig:estim-error-pd}
    \end{subfigure}
    \begin{subfigure}[b]{0.49\textwidth}
        \centering
        \includegraphics[width=\textwidth]{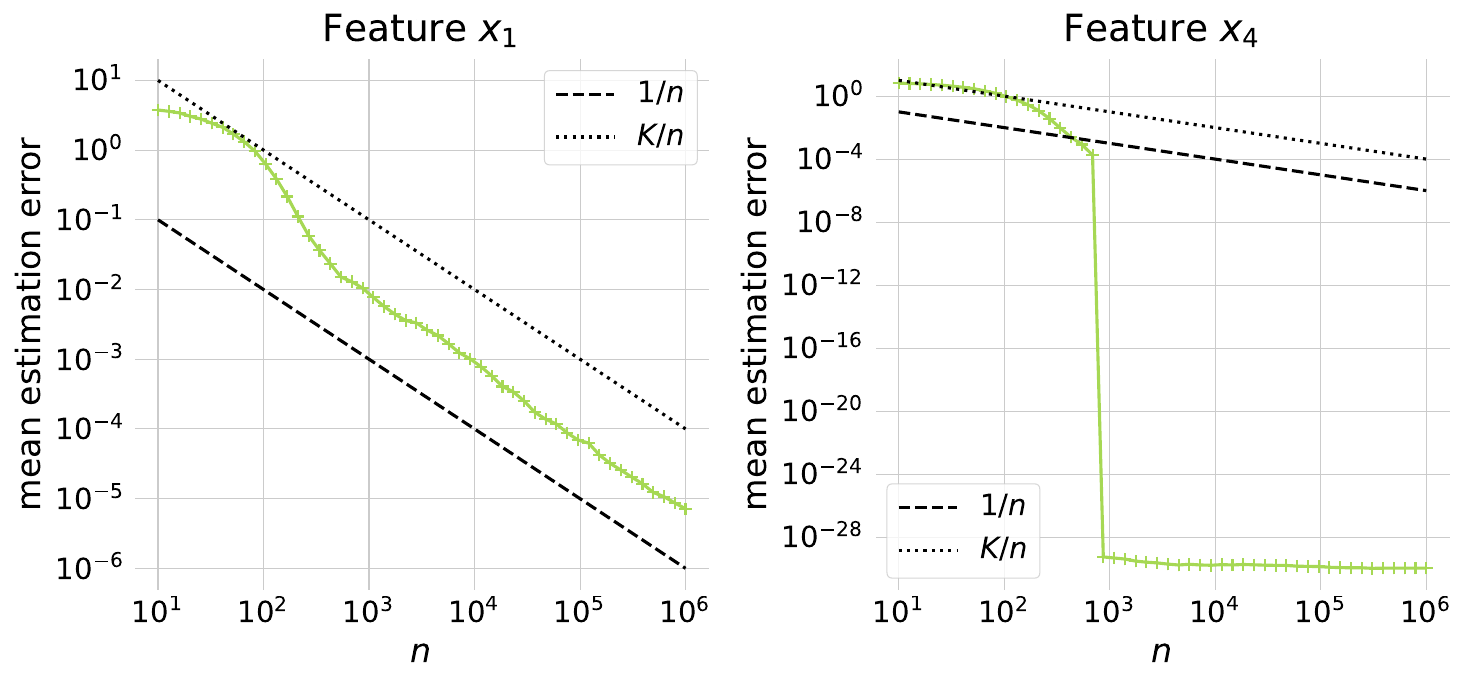}
        \caption{Mean estimation error for \ac{ALE}}
        \label{fig:estim-error-ale}
    \end{subfigure}
    \caption{Mean estimation errors on \textit{Friedman1} for $X_1$ and $X_4$. For each sample size $n$, the variances are averaged over all grid points. Both axes are log-scale.}
    \label{fig:mc-variance}
\end{figure*}

Our results for the effect of sample size are shown in Fig.~\ref{fig:mc-variance} for \textit{Friedman1}.
For \ac{PD} on holdout data, we know that there is no estimation bias (cf.~\S\ref{sec:theory-pd}) and Fig.~\ref{fig:estim-error-pd} contains only estimation variance.
We observe the expected polynomial decrease at roughly $1/n$ for features with interactions ($X_1$) and a negligible error across sample sizes for features without interactions ($X_4$), as expected for centered \ac{PD}.

For \ac{ALE}, we know from \S\ref{sec:theory-ale} that there are also estimation biases in addition to variance due to discretization and when $\nsk=0$.
With interactions ($X_1$), the observed estimation error in Fig.~\ref{fig:estim-error-ale} is closer to $\Kbins/\nsamples$ for small sample sizes but gets closer to $1/\nsamples$ as the sample size increases.
For features without interactions, we observe a sharp drop in the estimation error at $n=K$, reducing it to the expected negligible level.
This may be due to reduced estimation bias at this sample size, as each interval could, in principle, now contain at least one sample.

%
%
%
%%%%%%%%%%%%%%%%%%%%%%%%%%%%%%%%%%%%%%%%%%%%%%%%%%%%%%%%%%%%%%%%%%
%
%                         CONCLUSION
%
%%%%%%%%%%%%%%%%%%%%%%%%%%%%%%%%%%%%%%%%%%%%%%%%%%%%%%%%%%%%%%%%%%
\section{Conclusion \& Future Work}\label{sec:conclusion}

\textbf{Summary.}
In this work, we presented an estimator-level analysis of global feature effect estimation for \ac{PD} and \ac{ALE}. 
We derived a full \ac{MSE} decomposition that separates model bias, estimation bias, model variance, and estimation variance, and we analyzed these components theoretically.
Our results show that the model bias component follows directly from systematic biases in the model $\fhat$ for \ac{PD} or in the finite differences of $\fhat$ for \ac{ALE}.
For \ac{PD}, the estimation bias is zero on holdout data. 
For \ac{ALE}, it consists of a bias when $\nsk=0$, and a discretization bias due to binning.
For both \ac{PD} and \ac{ALE}, we derived upper bounds on the model variance in terms of pointwise variance of $\fhat$ (\ac{PD}) or its finite differences (\ac{ALE}).
Estimation variance is governed by (1) the sample size, scaling as $1/\nsamples$ for \ac{PD} and as the expected inverse bin counts for \ac{ALE}, explaining \ac{ALE}’s sensitivity in small-sample regimes, and (2) the variance of the \ac{ICE} curves / finite differences. For centered \ac{PD} and generally for \ac{ALE}, the latter depends only on interactions with $\XS$.

We validated our theoretical findings through simulations by comparing feature effect estimation on training and validation data and a \ac{CV}-based strategy. 
Our empirical results showed that potential bias from estimating feature effects on the training data is negligible across our settings, including when models overfit (RQ1).
Differences are instead dominated by sample-size effects: validation-set-based estimation yields higher variance (for \ac{ALE} also higher bias), while training-set- and \ac{CV}-based estimation attain the lowest \ac{MSE}.
Variance decomposition (RQ2) shows that the estimation variance is consistently highest on the smaller validation set and more pronounced for \ac{ALE}, in line with our theoretical results. \ac{CV} reduces model variance by averaging across fits (particularly beneficial for overfitting models) and estimation variance through a higher effective sample size compared to a single validation set. The observed estimation error (RQ3) confirms our theoretical results on sample size and interaction effects.

\textbf{Practical implications.}
While using holdout data is theoretically cleaner, our results indicate that training-set-based feature effect estimation is empirically safe and often preferable because it has a larger sample size. 
\ac{CV}-based estimation emerges as a robust alternative, particularly for overfitting models. Although such models are typically identified via appropriate performance estimation and ruled out, this may not always be perfectly possible in all applications, in which cases \ac{CV}-based feature effect estimation can be a safer option.

\textbf{Limitations \& future work.}
Our empirical results are themselves estimates with errors from finite repetitions, which can lead to artifacts, such as estimation variance estimates exceeding total variance (cf. Tab. \ref{tab:res-ablation-f1}). Additionally, the empirical analysis considers only low-dimensional settings and two model classes.
% Second, our analysis assumed that we want to be ``true to the data''. While theoretical errors w.r.t. a model can be derived from our results, an empirical analysis w.r.t. the true model feature effect is missing.
Further, our theoretical analysis leaves room for future research, including: (1)~a tighter bias analysis for cases in which $\DatasetRV\perp \fhat$ does not hold to formalize what happens for training-set-based estimation, (2)~a formal theory for \ac{CV}-based estimation results, (3)~extensions of our analysis to distribution shifts between training and estimation data, which is another practically relevant issue.
% (4)~Finally, extending the analysis to higher-dimensional feature effects and connecting it to interactions would generalize our findings.

%%%%%%%%%%%%%%%%%%%%%%%%%%%%%%%%%%%%%%%%%%%%%%%%%%%%%%%%%%%%%%%%%%
%%%%%%%%%%%%%%%%%%%%%%%%%%%%%%%%%%%%%%%%%%%%%%%%%%%%%%%%%%%%%%%%%%
% \begin{credits}
% %\subsubsection{\ackname} A bold run-in heading in small font size at the end of the paper is used for general acknowledgments, for example: This study was funded by X (grant number Y).

% \subsubsection{\discintname}
% The authors have no competing interests to declare that are relevant to the content of this article.
% \end{credits}
\bibliographystyle{splncs04}
\bibliography{mybibliography}
%
% \clearpage
%%%%%%%%%%%%%%%%%%%%%%%%%%%%%%%%%%%%%%%%%%%%%%%%%%%%%%%%%%%%%%%%%%
%%%%%%%%%%%%%%%%%%%%%%%%%%%%%%%%%%%%%%%%%%%%%%%%%%%%%%%%%%%%%%%%%%

%%%%%%%%%%%%%%%%%%%%%%%%%%%%%%%%%%%%%%%%%%%%%%%%%%%%%%%%%%%%%%%%%%
%
%                           APPENDIX
%
%%%%%%%%%%%%%%%%%%%%%%%%%%%%%%%%%%%%%%%%%%%%%%%%%%%%%%%%%%%%%%%%%%
\appendix
\section*{Appendix}

\section{Theoretical Evidence}

\subsection{Assumptions}\label{app:assumptions}

\begin{enumerate}[label=(\roman*),itemsep=4pt]

    \item\label{ass:FEs-Def-indep}
    \textit{Independence of feature effect definition and model.}
    Feature effects (both PD and ALE) are defined using expectations independent of the randomness in the training of $\fhat\sim \Pmodel$, meaning $\XnotS \perp \fhat$ for the theoretical PD / ALE.
    
    % \item\label{ass:holdout-indep}
    % \textit{Independence of estimation data and model}
    % % (needed for \S\ref{app:proof-pd-bias} to \S\ref{app:proof-ale-estimation-variance}, but not \S\ref{app:proof-decomposition}).
    % (only needed for some of our results, not for all, and is referenced accordingly).
    % The data used to estimate feature effects (e.g., via Monte Carlo integration for \ac{PD} or sample averages for \ac{ALE}) are drawn independently of $\fhat\sim \Pmodel$, in other words $\DatasetRV \perp \fhat$.
    % This corresponds to estimating feature effects on holdout data; estimation on training data may violate this independence.
    
    \item\label{ass:integrability} \textit{Integrability of $\fhat$ and $\ftrue$.}
    For all features $\XS$ and points $\xS$, the random variable $\fhat(\xS,\XnotS)$ is integrable and square-integrable under the joint law $\mathbb P_{(F, \XnotS)}$, i.e.,
    $\E_{(F, \XnotS)}[|\fhat(\xS,\XnotS)|]<\infty$ and $\E_{(F, \XnotS)}[\fhat(\xS,\XnotS)^2]<\infty$.
    In particular, this implies that for $\Pmodel$-almost all realizations of $\fhat$ we have $\E_{\XnotS \mid \fhat}[\fhat(\xS,\XnotS)^2]<\infty$ by Tonelli's theorem for conditional probabilities / stochastic kernels (since $\fhat(\xS,\XnotS)^2\ge 0$). Likewise $\E_{\XnotS \mid \fhat}[|\fhat(\xS,\XnotS)|]<\infty$.
    Analogously, $\ftrue(\xS,\XnotS)$ is integrable and square-integrable under $\mathbb P_{\XnotS}$, i.e.,
    $\E_{\XnotS}[|\ftrue(\xS,\XnotS)|]<\infty$ and $\E_{\XnotS}[\ftrue(\xS,\XnotS)^2]<\infty$.
    These conditions hold in particular for the special case that $\XnotS$ and $\fhat$ are independent, e.g., under Assumption~\ref{ass:FEs-Def-indep}.% or~\ref{ass:holdout-indep}.
    They ensure that the relevant expectations and variances are well defined (i.p., PD, ALE, and their estimators) and justify interchanging the order of integration (Fubini/Tonelli) where needed.
    
    \item\label{ass:fin-cond-exp}
    \textit{Finite conditional probabilities.}
    For all bins $\Isk$ considered,
    $
    \mathbb P(\XS \in \Isk) > 0,
    $
    and the conditional law
    $\mathbb P_{\XnotS\mid \XS \in \Isk}$
    is well-defined.
    
    \item\label{ass:bounded-tv}
    \textit{Uniform bounded total variation in $x_S$.}
    For all features $X_S$, there exists a nonnegative random variable $V(\fhat)$ such that, with probability 1 over $\fhat\sim\Pmodel$,
    \[
    \operatorname*{ess\,sup}_{\xnotS}
    \operatorname{TV}\!\big(t\mapsto \fhat(t,\xnotS)\big)
    \le V(\fhat),
    \qquad
    \E_F[V(\fhat)]<\infty.
    \]

    \item\label{ass:delta-moments}
    \textit{Integrability of finite differences.}
    For all features $\XS$, all $K$ and all $k$ considered, the finite differences of $\fhat$ are integrable and square-integrable w.r.t. the joint law given the $k$-th bin, i.e., under $\mathbb P_{(F, X) \mid \XS\in \IsKk}$.
    In other words, 
    $\E_{(F, X) \mid \XS\in \IsKk}[\,\FinDiffSK{\fhat}(k,\XnotS)^2]<\infty$, and analogously to~\ref{ass:integrability} this implies $\E_{\XnotS \mid \XS\in \IsKk,\fhat}[\,\FinDiffSK{\fhat}(k,\XnotS)^2]<\infty$ for $\Pmodel$-almost all $\fhat$.
    For integrability, we likewise have $\E_{(F, X) \mid \XS\in \IsKk}[\,|\FinDiffSK{\fhat}(k,\XnotS)|]<\infty$.
    Note that this follows from \ref{ass:bounded-tv} plus the square-integrability condition in \ref{ass:integrability}.
    As in~\ref{ass:integrability}, the same integrability conditions hold for $\FinDiffSK{\ftrue}(k,\XnotS)$ (with $\ftrue$ in place of $\fhat$).

    \item\label{ass:ale-l2}
    \textit{Existence and square-integrability of \ac{ALE} targets.}
    For all features $\XS$ and $\func \in \{\fhat,\ftrue\}$, the theoretical uncentered \ac{ALE} $\uALETheo{\func}(\xS)$ (Eq.~\eqref{eq:ale-theoretical-uncentered}) exists for all $\xS$.
    Moreover, the centered theoretical \ac{ALE} $\cALETheo{\func}(\xS)$ exists and has finite second moment w.r.t.\ $\XS$, i.e.,
    \(
    \E_{\XS}\!\left[\cALETheo{\func}(\XS)^2\right] < \infty.
    \)
\end{enumerate}

\subsection{Bias-Variance-Decomposition of the Estimators (Eq.~(\ref{eq:full-pd-decomposition}))}\label{app:proof-decomposition}

\begin{proof}
    Fix a feature index $S$ and an evaluation point $\xS$. Let $\DatasetRV$ denote a random dataset used to estimate the feature effect.
    For better readability, we omit the point $\xS$.
    Under Assumption~\ref{ass:integrability} for either dependent or independent $\DatasetRV$ and $\fhat$,
    \begin{align*}
        &\E_{F}\E_{\DatasetRV\mid\fhat}\big[{(\PDTheo{\ftrue} - \PDEst{\fhat})}^2\big]
        = \E_{F} \E_{\DatasetRV\mid\fhat}[\PDTheo{\ftrue}^2 - 2\PDTheo{\ftrue}\PDEst{\fhat} + \PDEst{\fhat}^2] \\
        &= \PDTheo{\ftrue}^2 - 2\PDTheo{\ftrue}\E_{F}\E_{\DatasetRV\mid\fhat}[\PDEst{\fhat}] + \E_{F}\E_{\DatasetRV\mid\fhat}[\PDEst{\fhat}^2] \\
        &= \PDTheo{\ftrue}^2 - 2\PDTheo{\ftrue}\E_{F}\E_{\DatasetRV\mid\fhat}[\PDEst{\fhat}] + \E_{F}\Var_{\DatasetRV\mid\fhat}[\PDEst{\fhat}]  + \E_{F}[\E_{\DatasetRV\mid\fhat}{[\PDEst{\fhat}]}^2] \\
        &= \PDTheo{\ftrue}^2 - 2\PDTheo{\ftrue}\E_{F}\E_{\DatasetRV\mid\fhat}[\PDEst{\fhat}] + \E_{F}\Var_{\DatasetRV\mid\fhat}[\PDEst{\fhat}] %\\
        %& \quad 
        + \Var_{F}\E_{\DatasetRV\mid\fhat}[\PDEst{\fhat}] 
        \\ & \quad + \E_{F}[\E_{\DatasetRV\mid\fhat}[\PDEst{\fhat}]]^2 \\
        &= {(\PDTheo{\ftrue} - \E_{F}\E_{\DatasetRV\mid\fhat}[\PDEst{\fhat}])}^2 + \Var_{F}\E_{\DatasetRV\mid\fhat}[\PDEst{\fhat}] + \E_{F}\Var_{\DatasetRV\mid\fhat}[\PDEst{\fhat}].
    \end{align*}
    The proof for \ac{ALE} follows by replacing $\PDTheo{\ftrue}$ with $\cALETheo{\ftrue}$ and $\PDEst{\fhat}$ with $\cALEEst{\fhat}$ in the equations above.
    For this replacement argument, it suffices that all involved quantities are well-defined (Assumption~\ref{ass:ale-l2}) and have finite second moments (Assumption~\ref{ass:delta-moments}), and that $\ftrue$, and hence $\cALETheo{\ftrue}$, is non-random w.r.t. $\Pmodel$.
    \qed
\end{proof}

\subsection{Model Bias of the \ac{PD} (Eq.~(\ref{eq:pd-model-bias}))}\label{app:proof-pd-bias}

\begin{proof}
Fix $\xS$.
Consider the product probability space $(\fhat,\XnotS)\sim \Pmodel\otimes \mathbb P_{\XnotS}$ (cf.~Assumption\ref{ass:FEs-Def-indep}).
By Assumption~\ref{ass:integrability}, $\fhat(\xS,\XnotS)$ is integrable under $\Pmodel\otimes \mathbb P_{\XnotS}$, hence $\PDTheo{\fhat}(\xS)=\E_{\XnotS}[\fhat(\xS,\XnotS)]$ exists for $\Pmodel$-almost all realizations of $\fhat$ and is integrable w.r.t.\ $\Pmodel$.
Now $\E_{F}\!\big[\PDTheo{\fhat}(\xS)\big] - \PDTheo{\ftrue}(\xS)$ is equal to
\begin{align*}
    \E_{F}\!\left[
    \E_{\XnotS}\!\big[\fhat(\xS,\XnotS)\big]\right] - \E_{\XnotS}\!\big[\ftrue(\xS,\XnotS)\big] %\\
    % &=
    % \int_{\mathcal F}
    % \int_{\mathcal X_{\backslash S}}
    % \fhat(\xS,\xnotS)\,
    % \mathrm d\mathbb P_{\XnotS}(\xnotS)\,
    % \mathrm d\Pmodel(\fhat)
    % -
    % \int_{\mathcal X_{\backslash S}}
    % \ftrue(\xS,\xnotS)\,
    % \mathrm d\mathbb P_{\XnotS}(\xnotS) \\
    % &=
    % \int_{\mathcal X_{\backslash S}}
    % \int_{\mathcal F}
    % \fhat(\xS,\xnotS)\,
    % \mathrm d\Pmodel(\fhat)\,
    % \mathrm d\mathbb P_{\XnotS}(\xnotS)
    % -
    % \int_{\mathcal X_{\backslash S}}
    % \ftrue(\xS,\xnotS)\,
    % \mathrm d\mathbb P_{\XnotS}(\xnotS) \\
    %&=
    = \E_{\XnotS}\!\left[
    \E_{F}\!\big[\fhat(\xS,\XnotS)\big]
    - \ftrue(\xS,\XnotS)
    \right].
\end{align*}
This follows by an application of Fubini's theorem (exchange of integrals) since $\E_{F}\E_{\XnotS}[|\fhat(\xS,\XnotS)|]<\infty$ by Assumption~\ref{ass:integrability}.
\qed
\end{proof}

\subsection{Model Variance of the \ac{PD} (Eq.~(\ref{eq:pd-model-variance}))}\label{app:proof-pd-theoretical-variance}

\begin{proof}
As in \S\ref{app:proof-pd-bias}, fix $\xS$ and suppose Assumptions~\ref{ass:FEs-Def-indep}~and~\ref{ass:integrability} hold.
Then, the variance of the theoretical model \ac{PD} exists and is:
\begin{equation*}
    \Var_{F}\big[\PDTheo{\fhat}(\xS)\big]
    =
    \E_{F}\Big[
    \big(\E_{\XnotS}[\fhat(\xS,\XnotS)]
    -\E_{F}\E_{\XnotS}[\fhat(\xS,\XnotS)]\big)^2
    \Big].
\end{equation*}
Applying Fubini's theorem as $\E_{F}\E_{\XnotS}[|\fhat(\xS,\XnotS)|] < \infty$ by Assumption~\ref{ass:integrability}, we have
    $\E_{F}\E_{\XnotS}[\fhat(\xS,\XnotS)] = \E_{\XnotS}\E_{F}[\fhat(\xS,\XnotS)].$
By linearity of expectations,
$
    \Var_{F}\big[\PDTheo{\fhat}(\xS)\big]
    =
    \E_{F}\Big[
    \big(\E_{\XnotS}\big[\fhat(\xS,\XnotS)
    -\E_{F}[\fhat(\xS,\XnotS)]\big]\big)^2
    \Big].
$
Defining $U:=\fhat(\xS,\XnotS)-\E_{F}[\fhat(\xS,\XnotS)]$ and applying Jensen's inequality 
to $\phi(t)=t^2$
%to the convex function $\phi(t)=t^2$
yields:
$
\big(\E_{\XnotS}[U]\big)^2 \le \E_{\XnotS}[U^2].
$
Therefore,
$
    \Var_{F}\big[\PDTheo{\fhat}(\xS)\big]
    \le
    \E_{F}\E_{\XnotS}[U^2].
$
Since $U^2\ge 0$, Tonelli's theorem completes the proof: 

$
    \Var_{F}\big[\PDTheo{\fhat}(\xS)\big]
    \le
    \E_{F}\E_{\XnotS}[U^2]
    =
    \E_{\XnotS}\E_{F}[U^2]
    =
    \E_{\XnotS}\Var_{F}\big[\fhat(\xS,\XnotS)\big].
$
\qed
\end{proof}

\subsection{Estimation Variance of the \ac{PD} (Eq.~(\ref{eq:pd-estimation-variance}))}\label{app:proof-pd-mc-variance}

\begin{proof}
Fix $\xS$ and a model $\fhat$.
Suppose Assumption~\ref{ass:integrability} holds,
and consider i.i.d.\ draws $D=\{\XnotS^{(i)}\}_{i=1}^{\nsamples}$ from $\mathbb P_{\XnotS \mid \fhat}$, conditional on $\fhat$.
Define $\hat Y_S^{(i)} := \fhat(\xS,\XnotS^{(i)})$, then by construction the $\hat Y_S^{(i)}$ are i.i.d.\ conditional on $\fhat$,
and by Assumption~\ref{ass:integrability}, we have $\E_{\XnotS^{(i)} \mid \fhat}\!\big[(\hat Y_S^{(i)})^2\big]<\infty$ for $\Pmodel$-almost all $\fhat$.
Thereby
\begin{align*}
    &\Var_{\DatasetRV\mid\fhat}\bigl[\PDEst{\fhat}(\xS)\bigr]
    =\Var_{\DatasetRV\mid\fhat}\!\left[\frac{1}{\nsamples}\sum_{i=1}^{\nsamples} \hat Y_S^{(i)}\right]
    =\frac{1}{\nsamples^2}\Var_{\DatasetRV\mid\fhat}\!\left[\sum_{i=1}^{\nsamples} \hat Y_S^{(i)}\right] \\
    % For the next step, independence of $X^{(i)}$ conditional on $\fhat$ is important
    &=\frac{1}{\nsamples^2}\left(
    \sum_{i=1}^{\nsamples} \Var_{\DatasetRV\mid\fhat}[\hat Y_S^{(i)}]
    \right)
    =\frac{1}{\nsamples^2}\left(\nsamples\,\Var_{\DatasetRV\mid\fhat}[\hat Y_S^{(1)}]\right)
    =\frac{1}{\nsamples}\Var_{\DatasetRV\mid\fhat}\!\bigl[\fhat(\xS,\XnotS)\bigr],
\end{align*}
with $\XnotS$ in the last step being drawn conditional on $\smash{\fhat}$, so from $\smash{\DatasetRV\mid\fhat}$.
Taking expectation over the training randomness of $\fhat$ gives the proof.
\qed
\end{proof}

\subsection{\ac{ALE} Estimator Bias w.r.t. Binned Population \ac{ALE} (\S\ref{sec:theory-ale})}\label{app:proof-ale-estimation-bias}

\begin{proof}
Fix $\xS$, bins $\{\Isk\}_{\kbin=1}^{\Kbins}$, and a realization of $\fhat$.
Suppose Assumptions \ref{ass:fin-cond-exp} and \ref{ass:delta-moments} hold. Moreover, assume $D=\{\X^{(i)}\}_{i=1}^{\nsamples}$ (used for estimation) are i.i.d.\ draws from $\mathbb P_{\X}$, independent of $\fhat$, and assume $\nsk>0$ for all bins $k\le \ksx$. % (so that $\uALEEst{\fhat}(\xS)$ is well-defined).
Let $\BBinIndex=(\BBinIndex^{(1)},\ldots,\BBinIndex^{(\nsamples)})$ be the bin assignments of the samples, where $\BBinIndex^{(i)}=k_S(\XS^{(i)})$.
Then, conditional on the \emph{full} assignment vector $\BBinIndex$, i.e., on the product event $\bigcap_{i=1}^n\{\XS^{(i)}\in I_S({\BBinIndex^{(i)}})\}$, the samples remain independent; in particular, for each $\kbin$ the subcollection $\{\XnotS^{(i)} : \BBinIndex^{(i)}=\kbin\}$ is i.i.d.\ with $\XnotS^{(i)} \sim \mathbb P_{\XnotS\mid \XS\in\Isk}$.
Therefore, using the law of total expectation and $\DatasetRV\perp\fhat$, we obtain
\begin{align*}
    &\E_{\DatasetRV\mid\fhat}\!\left[\uALEEst{\fhat}(\xS)\right]
    =
    \E_{\BBinIndex}\E_{\DatasetRV \mid \BBinIndex}\!\bigg[
    \sum_{\kbin=1}^{\ksx}\frac{1}{\nsk}\sum_{i:\XS^{(i)}\in\Isk}
    \FinDiff{\fhat}(\kbin,\XnotS^{(i)})
    \bigg] \\
    &=
    \E_{\BBinIndex}\!\bigg[
    \sum_{\kbin=1}^{\ksx}\frac{1}{\nsk}\sum_{i:\XS^{(i)}\in\Isk}
    \E_{\DatasetRV \mid \BBinIndex}\!\left[\FinDiff{\fhat}(\kbin,\XnotS^{(i)})\right]
    \bigg] \\
    &=
    \E_{\BBinIndex}\!\bigg[
    \sum_{\kbin=1}^{\ksx}
    \E_{\XnotS\mid\XS\in\Isk}\!\left[\FinDiff{\fhat}(\kbin,\XnotS)\right]
    \bigg] =
    \E_{\BBinIndex}\!\big[\uALETheoBinned{\fhat}(\xS)\big]
    = \uALETheoBinned{\fhat}(\xS).
\end{align*}
Thus, $\uALEEst{\fhat}$ is unbiased for $\uALETheoBinned{\fhat}$: $\E_{\DatasetRV\mid\fhat}\!\big[\uALEEst{\fhat}(\xS)\big]-\uALETheoBinned{\fhat}(\xS) = 0$.
\qed
\end{proof}

\subsection{Model Bias of the \ac{ALE} (\S\ref{sec:theory-ale})}\label{app:proof-ale-model-bias}

\begin{proof}
Fix $\xS$ and 
let $\fhat\sim \Pmodel$.
Suppose Assumptions~\ref{ass:FEs-Def-indep}, \ref{ass:fin-cond-exp}, \ref{ass:bounded-tv}, and \ref{ass:delta-moments} hold, so $(\XS,\XnotS)\sim \mathbb P_X$ is an independent population draw, i.e., $(\XS,\XnotS)\perp \fhat$, and $\uALETheo{\fhat}(\xS)$ and $\uALETheo{\ftrue}(\xS)$ exist by Assumption~\ref{ass:ale-l2}.
For $\Kbins\in\mathbb N$, define
$
g_{\Kbins}(\fhat)
:=
\sum_{\kbin=1}^{\ksKx}
\E_{\XnotS \mid \XS \in \IsKk}\big[\FinDiffSK{\fhat}(\kbin,\XnotS)\big],
$
so that $\uALETheo{\fhat}(\xS)=\lim_{\Kbins\to\infty} g_{\Kbins}(\fhat)$ by definition.
Then,
\begin{align*}
\E_{F}\!\left[\uALETheo{\fhat}(\xS)\right]
&=
\int_{\mathcal F}
\lim_{\Kbins\to\infty} g_{\Kbins}(\fhat)\,
\mathrm d\Pmodel(\fhat)
=
\lim_{\Kbins\to\infty}
\int_{\mathcal F}
g_{\Kbins}(\fhat)\,
\mathrm d\Pmodel(\fhat)
\\[0.5em]
&=
\lim_{\Kbins\to\infty}
\sum_{\kbin=1}^{\ksKx}
\int_{\mathcal F}
\E_{\XnotS \mid \XS \in \IsKk}\!\left[\FinDiffSK{\fhat}(\kbin,\XnotS)\right]
\,\mathrm d\Pmodel(\fhat)
\\[0.5em]
&=
\lim_{\Kbins\to\infty}
\sum_{\kbin=1}^{\ksKx}
\E_{\XnotS \mid \XS \in \IsKk}\!\left[
\E_{F}\!\left[\FinDiffSK{\fhat}(\kbin,\XnotS)\right]
\right].
\end{align*}
The second equality follows from the Dominated Convergence Theorem.
Indeed, $g_{\Kbins}(\fhat)\to \uALETheo{\fhat}(\xS)$ pointwise in $\fhat$ by definition, and for every $\Kbins$,
\begin{align*}
\big|g_{\Kbins}(\fhat)\big|
&\le
\sum_{\kbin=1}^{\ksKx}
\E_{\XnotS \mid \XS \in \IsKk}\!\left[\big|\FinDiffSK{\fhat}(\kbin,\XnotS)\big|\right]
\le
\sum_{\kbin=1}^{\ksKx}
\operatorname*{ess\,sup}_{\xnotS}\big|\FinDiffSK{\fhat}(\kbin,\xnotS)\big|
\\
&\le
\operatorname*{ess\,sup}_{\xnotS}\!
\sum_{\kbin=1}^{\ksKx}\!\!
\big|\fhat(\zsKupper,\xnotS)-\fhat(\zsKlower,\xnotS)\big|\\
&\le
\operatorname*{ess\,sup}_{\xnotS}
\operatorname{TV}\!\big(t\mapsto \fhat(t,\xnotS)\big)
\le
V(\fhat),
\end{align*}
where $V(\fhat)$ is the integrable bound from Assumption~\ref{ass:bounded-tv}.
The last equality above follows from Fubini's theorem, using Assumption~\ref{ass:delta-moments} (integrability) for $\mathbb P_{\XnotS\mid \XS\in\IsKk,\fhat}=\mathbb P_{\XnotS\mid \XS\in\IsKk}$ since $(\XS,\XnotS)\perp \fhat$.
By linearity of expectation,
\begin{align*}
&\E_{F}\!\left[
\uALETheo{\fhat}(\xS)
\right]
-
\uALETheo{\ftrue}(\xS)
=\\
&\lim_{\Kbins\to\infty}
\sum_{\kbin=1}^{\ksKx}
\E_{\XnotS\mid \XS \in \IsKk}
\!\left[
\E_{F}\!\left[
\FinDiffSK{\fhat}(\kbin,\XnotS)
\right]
-
\FinDiffSK{\ftrue}(\kbin,\XnotS)
\right].
\end{align*}
Thus, this bias term vanishes if the finite differences of $\fhat$ are unbiased w.r.t.~those of $\ftrue$.
One sufficient condition is that $\fhat$ is a pointwise unbiased estimator of $\ftrue$.
% \[
% \mathbb E_F\!\left[
% \Delta^{(K)}_{\hat f,S}(k,x_{\backslash S})
% \right]
% =
% \mathbb E_F\!\left[
% \hat f(z^{(K)}_{k,S},x_{\backslash S})
% \right]
% -
% \mathbb E_F\!\left[
% \hat f(z^{(K)}_{k-1,S},x_{\backslash S})
% \right]
% =
% f(z^{(K)}_{k,S},x_{\backslash S})
% -
% f(z^{(K)}_{k-1,S},x_{\backslash S}),
% \]
\qed
\end{proof}

\subsection{Model Variance of the \ac{ALE} (Eq.~(\ref{eq:ale-model-variance-bound}))}\label{app:proof-ale-model-variance}

\begin{proof}
Fix $\xS$,
let $\fhat\sim \Pmodel$,
and suppose Assumptions~\ref{ass:FEs-Def-indep}, ~\ref{ass:fin-cond-exp}, and~\ref{ass:delta-moments} hold,
so that as in \S\ref{app:proof-ale-model-bias}, for each bin index $\kbin$, all expectations w.r.t.\ $\mathbb P_{\XnotS\mid \EventInI}$ are independent of $\fhat$.
For readability, we abbreviate the event $\{\XS \in \Isk\}$ by $\EventInI$.
Then:

\begin{align*}
&\Var_{F}\!\left[\uALETheoBinned{\fhat}(\xS)\right]
=
\Var_{F}\!\left(
\sum_{\kbin=1}^{\ksx}
\E_{\XnotS \mid \EventInI}
\big[
\FinDiff{\fhat}(\kbin, \XnotS)
\big]
\right) \\
&=
\E_{F}\!\bigg[
\bigg(
\sum_{\kbin=1}^{\ksx}
\E_{\XnotS \mid \EventInI}
\big[
\FinDiff{\fhat}(\kbin, \XnotS)
\big]
-
\E_{F}\!\bigg[
\sum_{\kbin=1}^{\ksx}
\E_{\XnotS \mid \EventInI}
\big[
\FinDiff{\fhat}(\kbin, \XnotS)
\big]
\bigg]
\bigg)^2
\bigg] \\
&=
\E_{F}\!\bigg[
\bigg(
\sum_{\kbin=1}^{\ksx}
\E_{\XnotS \mid \EventInI}
\bigg[
\FinDiff{\fhat}(\kbin, \XnotS)
-
\E_{F}
\big[
\FinDiff{\fhat}(\kbin, \XnotS)
\big]
\bigg]
\bigg)^2
\bigg].
\end{align*}
The last line follows from the linearity of expectations and exchanging
$\E_{F}$ and
$\E_{\XnotS\mid \EventInI}$
by an application of Fubini's theorem.
Indeed, using Assumption~\ref{ass:delta-moments} for $\fhat \perp (\XS, \XnotS)$ ensures that the resulting joint integrand is integrable.
Define
$
U_k
:=
\E_{\XnotS \mid \EventInI}
\!\big[
\FinDiff{\fhat}(\kbin, \XnotS)
-
\E_{F}
\big[
\FinDiff{\fhat}(\kbin, \XnotS)
\big]
\big],
$
and let
$
\mathbf U
=
\big(
U_1,\ldots,U_{\ksx}
\big)^{\top}.
$
By the Cauchy-Schwarz inequality,
$
\big(
\sum_{\kbin=1}^{\ksx} U_k
\big)^2
=
\langle \mathbf 1, \mathbf U \rangle^2
\le
\lVert \mathbf 1 \rVert^2 \, \lVert \mathbf U \rVert^2
=
\ksx \sum_{\kbin=1}^{\ksx} U_k^2 .
$
Moreover,
$
\E_{F}
\!\big[
\ksx\!\sum_{\kbin=1}^{\ksx} U_k^2
\big]
=
\ksx \sum_{\kbin=1}^{\ksx}
\E_{F}\!\big[ U_k^2 \big]
$
by linearity of expectation.
By Assumption~\ref{ass:delta-moments}, applying Jensen's inequality to $\varphi(t)=t^2$, we obtain
$
U_k^2
% =
% \left(
% \E_{\XnotS\mid \EventInI}
% \!\left[
% \FinDiff{\fhat}(\kbin, \XnotS)
% -
% \E_{F}
% \big[
% \FinDiff{\fhat}(\kbin, \XnotS)
% \big]
% \right]
% \right)^2
% \]
% \[
\le
\E_{\XnotS\mid \EventInI}
\big[
\big(
\FinDiff{\fhat}(\kbin, \XnotS)
-
\E_{F}
[
\FinDiff{\fhat}(\kbin, \XnotS)
]
\big)^2
\big].$
Putting everything together and applying Tonelli's theorem (the integrand $U_k^2$ is nonnegative) gives the proof:

\begin{align*}
\Var_{F}
\!\bigg[
\uALETheoBinned{\fhat}(\xS)
\bigg]
&\le
\ksx\sum_{\kbin=1}^{\ksx}
\E_{F}
\E_{\XnotS\mid \EventInI}
\!\left[
\left(
\FinDiff{\fhat}(\kbin, \XnotS)
-
\E_{F}
\big[
\FinDiff{\fhat}(\kbin, \XnotS)
\big]
\right)^2
\right] \\
&=
\ksx\sum_{\kbin=1}^{\ksx}
\E_{\XnotS\mid \EventInI}
\E_{F}
\!\left[
\left(
\FinDiff{\fhat}(\kbin, \XnotS)
-
\E_{F}
\big[
\FinDiff{\fhat}(\kbin, \XnotS)
\big]
\right)^2
\right] \\
&=
\ksx\sum_{\kbin=1}^{\ksx}
\E_{\XnotS\mid \EventInI}
\Var_{F}
\!\left[
\FinDiff{\fhat}(\kbin, \XnotS)
\right].\tag*{\(\square\)}
\end{align*}
\end{proof}

\subsection{Estimation Variance of the \ac{ALE} (Eq.~(\ref{eq:ale-estimation-variance-cond}))}\label{app:proof-ale-estimation-variance}

\begin{proof}
Fix $\xS$ and bins $\Isk$, suppose Assumptions~\ref{ass:fin-cond-exp}, \ref{ass:delta-moments} and \ref{ass:ale-l2} hold, i.p. with $D=\{X^{(i)}\}_{i=1}^{\nsamples}$ being i.i.d.\ draws from $\mathbb P_{X \mid \fhat}$, conditional on $\fhat$,
and assume $\nsk>0 \ \forall k$.
For any fixed $k$, the variance of the finite differences $\sigma_k^2(\fhat):=\Var_{\XnotS\mid \XS\in \Isk, \fhat}[\FinDiff{\fhat}(k,\XnotS)]$ is finite by Assumption~\ref{ass:delta-moments}: $\sigma_k^2(\fhat)<\infty$.
Now, for each $k$, conditional on a model $\fhat$ and on the full bin assignment vector $\BBinIndex$ of all samples,
the variables $\{\FinDiff{\fhat}(k,\XnotS^{(i)}): \BBinIndex^{(i)}=k\}$ (all those falling into $\Isk$) are i.i.d.\ with variance $\sigma_k^2(\fhat)$.
Since also $\nsk$ and the events $\{X_S^{(i)}\in \Isk\}$ are fixed conditional on $\BBinIndex$ and $\fhat$ (similar to \S\ref{app:proof-ale-estimation-bias}),
% CB: The next step, this calculation of the variance, is actually a little bit intricate, because the summation terms themselves are random. I have calculated the proof on paper, is a little nasty / calculation-intensive, but not insightful or important.
this gives us $\Var_{\DatasetRV\mid\fhat,\BBinIndex}[\widehat{\mu}_k(\fhat)]=\sigma_k^2(\fhat)/\nsk$,
where we define the unaccumulated value of the $k$-th bin as
$
\widehat{\mu}_k(\fhat)
:=
\frac{1}{\nsk}\sum_{i:X_S^{(i)}\in \Isk}
\FinDiff{\fhat}\big(k,\XnotS^{(i)}\big).
$
Moreover, conditional on $\BBinIndex$ and $\fhat$, different bins use disjoint subsets of the i.i.d.\ sample, hence
$\mathrm{Cov}_{\DatasetRV\mid\fhat,\BBinIndex}[\widehat{\mu}_k(\fhat),\widehat{\mu}_\ell(\fhat)]=0$ for $k\neq \ell$.
Altogether we get:
\begin{align*}
    &\E_{F}\E_{\BBinIndex\mid\fhat}\Var_{\DatasetRV\mid\fhat,\BBinIndex}\!\left[\uALEEst{\fhat}(\xS)\right]
    =
    \E_{F}\E_{\BBinIndex\mid\fhat}\Var_{\DatasetRV\mid\fhat,\BBinIndex}\bigg[\sum_{k\le \ksx}\widehat{\mu}_k(\fhat)\bigg]\\
    &=
    \sum_{k\le \ksx} \E_{F}\E_{\BBinIndex\mid\fhat}\Var_{\DatasetRV\mid\fhat,\BBinIndex}[\widehat{\mu}_k(\fhat)]
    =
    \sum_{k\le \ksx} \E_{F}\E_{\BBinIndex\mid\fhat}\left[\frac{1}{\nsk}\sigma_k^2(\fhat)\right]
    .\tag*{\(\square\)}
\end{align*}
\end{proof}

\section{Simulation Details \& Results}

\subsection{Models, Hyperparameters, and Model Performances}\label{app:model-hps}

The XGBoost implementation from \textit{xgboost}~\cite{chen_xgboost_2016}, and the \ac{GAM} implementation from \textit{pyGAM}~\cite{serven_pygam_2018} were used.
For the \ac{GAM}, we considered the number of basis functions ($\in[5,50]$) and the penalization control parameter ($\in[0.001,1000]$, log-uniform) as tuning parameters. For XGBoost, we used the parameter spaces from~\cite{probst_tunability_2019} (confined to trees as base learners).

Hyperparameter configurations for the overfitting models (OF) were carefully hand-picked to achieve strong performance on the training data while performing relatively poorly on holdout data.
The optimal hyperparameters (OT) were selected by tuning the models on a separate data sample of size $n$ ($1250$ or $10000$) for 200 trials using a
Tree-structured Parzen Estimator (TPE)~\cite{bergstra2011algorithms}
with \ac{MSE} on separate holdout data ($10000$ samples for reliable performance estimates) as minimization objective.
For both OF and OT models, the final selected hyperparameter configurations can be found on Github (\href{https://github.com/slds-lmu/paper_2026_xai_feature_effects_code}{link}).

We evaluate model performance across all repetitions on both the training and test data, with $10000$ test samples to obtain reliable performance estimates.
As intended, overfitting models show better training performance than the optimally tuned models, but underperform on test data and exhibit higher variance in their generalization performance. A linear regression model is used as a baseline and is outperformed by all OT models. These performance metrics can be found on GitHub (\href{https://github.com/slds-lmu/paper_2026_xai_feature_effects_code}{link}).

\subsection{Further Results: Bias-Variance-Analysis}\label{app:results-bias-variance}

\begin{table}[h]
    \centering
    \scriptsize
    \caption{Results for \ac{PD} on \textit{Friedman1} averaged over 100 grid points. Bold numbers indicate the minimum per metric-feature-model-size-combination.}
   \begin{tabular}{lll|rrr|rrr|rrr}
   \toprule
     &  & \textbf{Feature} & \multicolumn{3}{c}{$\mathbf{x_1}$} & \multicolumn{3}{c}{$\mathbf{x_3}$} & \multicolumn{3}{c}{$\mathbf{x_5}$} \\
     &  & \textbf{Metric} & MSE & Bias & Var & MSE & Bias & Var & MSE & Bias & Var \\
     \midrule
\multirow[c]{12}{*}{\rotatebox{90}{$n=1250$}} & \multirow[c]{3}{*}{GAM\_OF} & train & 0.1184 & 0.0720 & 0.1173 & 0.1285 & \bfseries 0.0639 & 0.1289 & 0.1210 & \bfseries 0.0631 & 0.1212 \\
 &  & val & 0.1968 & 0.0929 & 0.1949 & 0.2084 & 0.0907 & 0.2073 & 0.2302 & 0.0955 & 0.2289 \\
 &  & CV & \bfseries 0.1062 & \bfseries 0.0697 & \bfseries 0.1049 & \bfseries 0.1094 & 0.0650 & \bfseries 0.1090 & \bfseries 0.1057 & 0.0652 & \bfseries 0.1051 \\
	\cmidrule(){2-12}
 & \multirow[c]{3}{*}{GAM\_OT} & train & 0.0166 & 0.0190 & 0.0168 & 0.0146 & 0.0238 & 0.0145 & 0.0145 & \bfseries 0.0166 & 0.0147 \\
 &  & val & 0.0264 & 0.0266 & 0.0266 & 0.0187 & 0.0236 & 0.0188 & 0.0199 & 0.0221 & 0.0201 \\
 &  & CV & \bfseries 0.0165 & \bfseries 0.0182 & \bfseries 0.0167 & \bfseries 0.0144 & \bfseries 0.0236 & \bfseries 0.0143 & \bfseries 0.0143 & 0.0166 & \bfseries 0.0146 \\
	\cmidrule(){2-12}
 & \multirow[c]{3}{*}{XGB\_OF} & train & 0.1361 & \bfseries 0.2297 & 0.0862 & \bfseries 0.4853 & \bfseries 0.6748 & 0.0310 & 0.1142 & \bfseries 0.2767 & 0.0390 \\
 &  & val & 0.1915 & 0.2885 & 0.1120 & 0.6402 & 0.7744 & 0.0419 & 0.1718 & 0.3097 & 0.0786 \\
 &  & CV & \bfseries 0.1196 & 0.2742 & \bfseries 0.0460 & 0.5429 & 0.7258 & \bfseries 0.0167 & \bfseries 0.1055 & 0.2853 & \bfseries 0.0249 \\
	\cmidrule(){2-12}
 & \multirow[c]{3}{*}{XGB\_OT} & train & \bfseries 0.0327 & \bfseries 0.1081 & 0.0217 & \bfseries 0.0393 & \bfseries 0.1511 & 0.0170 & \bfseries 0.0208 & \bfseries 0.0854 & 0.0140 \\
 &  & val & 0.0416 & 0.1206 & 0.0280 & 0.0509 & 0.1752 & 0.0209 & 0.0245 & 0.0925 & 0.0165 \\
 &  & CV & 0.0334 & 0.1227 & \bfseries 0.0190 & 0.0439 & 0.1718 & \bfseries 0.0149 & 0.0209 & 0.0941 & \bfseries 0.0125 \\
 \midrule
\multirow[c]{12}{*}{\rotatebox{90}{$n=10000$}} & \multirow[c]{3}{*}{GAM\_OF} & train & \bfseries 0.0094 & 0.0178 & \bfseries 0.0094 & \bfseries 0.0086 & \bfseries 0.0173 & \bfseries 0.0086 & \bfseries 0.0096 & 0.0192 & \bfseries 0.0096 \\
 &  & val & 0.0137 & 0.0212 & 0.0137 & 0.0117 & 0.0211 & 0.0117 & 0.0129 & 0.0235 & 0.0128 \\
 &  & CV & 0.0095 & \bfseries 0.0178 & 0.0095 & 0.0087 & 0.0173 & 0.0086 & 0.0097 & \bfseries 0.0189 & 0.0096 \\
	\cmidrule(){2-12}
 & \multirow[c]{3}{*}{GAM\_OT} & train & 0.0005 & 0.0082 & 0.0005 & \bfseries 0.0004 & 0.0024 & \bfseries 0.0004 & \bfseries 0.0004 & \bfseries 0.0035 & \bfseries 0.0004 \\
 &  & val & 0.0013 & \bfseries 0.0076 & 0.0013 & 0.0005 & \bfseries 0.0021 & 0.0005 & 0.0005 & 0.0049 & 0.0005 \\
 &  & CV & \bfseries 0.0005 & 0.0082 & \bfseries 0.0005 & 0.0004 & 0.0024 & 0.0004 & 0.0004 & 0.0036 & 0.0004 \\
	\cmidrule(){2-12}
 & \multirow[c]{3}{*}{XGB\_OF} & train & \bfseries 0.0232 & \bfseries 0.1194 & 0.0092 & \bfseries 0.0675 & \bfseries 0.2523 & 0.0040 & \bfseries 0.0302 & \bfseries 0.1594 & 0.0049 \\
 &  & val & 0.0326 & 0.1370 & 0.0143 & 0.0924 & 0.2949 & 0.0057 & 0.0386 & 0.1776 & 0.0073 \\
 &  & CV & 0.0235 & 0.1314 & \bfseries 0.0065 & 0.0860 & 0.2886 & \bfseries 0.0028 & 0.0344 & 0.1753 & \bfseries 0.0038 \\
	\cmidrule(){2-12}
 & \multirow[c]{3}{*}{XGB\_OT} & train & 0.0060 & \bfseries 0.0266 & 0.0055 & 0.0060 & \bfseries 0.0374 & 0.0047 & 0.0044 & \bfseries 0.0274 & 0.0038 \\
 &  & val & 0.0078 & 0.0318 & 0.0070 & 0.0073 & 0.0428 & 0.0057 & 0.0055 & 0.0327 & 0.0046 \\
 &  & CV & \bfseries 0.0052 & 0.0293 & \bfseries 0.0045 & \bfseries 0.0056 & 0.0427 & \bfseries 0.0039 & \bfseries 0.0041 & 0.0310 & \bfseries 0.0033 \\
	\bottomrule
	\end{tabular}
	\label{tab:res-pd-f1}
\end{table}

\begin{table}[h]
    \centering
    \scriptsize
    \caption{Results for \ac{ALE} on \textit{Friedman1} averaged over 100 grid points. Bold numbers indicate the minimum per metric-feature-model-size-combination.}
   \begin{tabular}{lll|rrr|rrr|rrr}
   \toprule
     &  & \textbf{Feature} & \multicolumn{3}{c}{$\mathbf{x_1}$} & \multicolumn{3}{c}{$\mathbf{x_3}$} & \multicolumn{3}{c}{$\mathbf{x_5}$} \\
     &  & \textbf{Metric} & MSE & Bias & Var & MSE & Bias & Var & MSE & Bias & Var \\
     \midrule
\multirow[c]{12}{*}{\rotatebox{90}{$n=1250$}} & \multirow[c]{3}{*}{GAM\_OF} & train & \bfseries 0.1280 & \bfseries 0.0809 & \bfseries 0.1258 & \bfseries 0.1290 & \bfseries 0.0629 & \bfseries 0.1295 & 0.1246 & \bfseries 0.0642 & 0.1248 \\
 &  & val & 0.4060 & 0.1738 & 0.3892 & 0.4199 & 0.2139 & 0.3875 & 0.4212 & 0.1265 & 0.4196 \\
 &  & CV & 0.1745 & 0.1855 & 0.1451 & 0.1497 & 0.1479 & 0.1324 & \bfseries 0.1212 & 0.0951 & \bfseries 0.1162 \\
	\cmidrule(){2-12}
 & \multirow[c]{3}{*}{GAM\_OT} & train & \bfseries 0.0206 & \bfseries 0.0336 & 0.0201 & \bfseries 0.0146 & \bfseries 0.0235 & 0.0145 & \bfseries 0.0144 & \bfseries 0.0165 & 0.0146 \\
 &  & val & 0.0795 & 0.1668 & 0.0535 & 0.0403 & 0.1219 & 0.0263 & 0.0316 & 0.1086 & 0.0205 \\
 &  & CV & 0.0400 & 0.1449 & \bfseries 0.0196 & 0.0273 & 0.1159 & \bfseries 0.0144 & 0.0257 & 0.1147 & \bfseries 0.0130 \\
	\cmidrule(){2-12}
 & \multirow[c]{3}{*}{XGB\_OF} & train & 3.6571 & 1.6329 & 1.0248 & 1.2235 & 0.8780 & 0.4683 & 2.6986 & 1.4763 & 0.5371 \\
 &  & val & 4.8262 & 0.5670 & 4.6600 & 1.5224 & \bfseries 0.7511 & 0.9914 & 1.9319 & 0.4747 & 1.7654 \\
 &  & CV & \bfseries 0.8781 & \bfseries 0.4926 & \bfseries 0.6573 & \bfseries 0.7944 & 0.7932 & \bfseries 0.1710 & \bfseries 0.5237 & \bfseries 0.4235 & \bfseries 0.3562 \\
	\cmidrule(){2-12}
 & \multirow[c]{3}{*}{XGB\_OT} & train & \bfseries 0.0591 & \bfseries 0.1629 & 0.0337 & \bfseries 0.0344 & \bfseries 0.0961 & 0.0260 & \bfseries 0.0482 & \bfseries 0.1701 & 0.0199 \\
 &  & val & 0.1411 & 0.2250 & 0.0937 & 0.1314 & 0.2818 & 0.0538 & 0.0770 & 0.1795 & 0.0463 \\
 &  & CV & 0.0878 & 0.2428 & \bfseries 0.0299 & 0.0920 & 0.2690 & \bfseries 0.0203 & 0.0563 & 0.1989 & \bfseries 0.0173 \\
 \midrule
\multirow[c]{12}{*}{\rotatebox{90}{$n=10000$}} & \multirow[c]{3}{*}{GAM\_OF} & train & \bfseries 0.0107 & 0.0272 & \bfseries 0.0103 & \bfseries 0.0087 & 0.0171 & \bfseries 0.0087 & \bfseries 0.0096 & 0.0191 & \bfseries 0.0096 \\
 &  & val & 0.0177 & 0.0312 & 0.0173 & 0.0127 & 0.0214 & 0.0127 & 0.0135 & 0.0237 & 0.0133 \\
 &  & CV & 0.0107 & \bfseries 0.0267 & 0.0103 & 0.0088 & \bfseries 0.0171 & 0.0088 & 0.0097 & \bfseries 0.0189 & 0.0097 \\
	\cmidrule(){2-12}
 & \multirow[c]{3}{*}{GAM\_OT} & train & \bfseries 0.0016 & 0.0239 & \bfseries 0.0011 & 0.0004 & 0.0023 & 0.0004 & \bfseries 0.0004 & \bfseries 0.0036 & \bfseries 0.0004 \\
 &  & val & 0.0036 & \bfseries 0.0233 & 0.0031 & 0.0005 & \bfseries 0.0023 & 0.0005 & 0.0005 & 0.0048 & 0.0005 \\
 &  & CV & 0.0016 & 0.0239 & 0.0011 & \bfseries 0.0004 & 0.0024 & \bfseries 0.0004 & 0.0004 & 0.0036 & 0.0004 \\
	\cmidrule(){2-12}
 & \multirow[c]{3}{*}{XGB\_OF} & train & 0.6796 & 0.8044 & 0.0336 & 0.4915 & 0.6802 & 0.0299 & 0.8473 & 0.9057 & 0.0279 \\
 &  & val & 0.2098 & 0.1518 & 0.1932 & 0.1711 & 0.3057 & 0.0803 & 0.1157 & \bfseries 0.1788 & 0.0866 \\
 &  & CV & \bfseries 0.0485 & \bfseries 0.1308 & \bfseries 0.0324 & \bfseries 0.1091 & \bfseries 0.3024 & \bfseries 0.0183 & \bfseries 0.0499 & 0.1821 & \bfseries 0.0173 \\
	\cmidrule(){2-12}
 & \multirow[c]{3}{*}{XGB\_OT} & train & 0.0070 & \bfseries 0.0252 & 0.0066 & \bfseries 0.0055 & \bfseries 0.0209 & 0.0052 & 0.0045 & \bfseries 0.0164 & 0.0044 \\
 &  & val & 0.0166 & 0.0392 & 0.0155 & 0.0112 & 0.0483 & 0.0091 & 0.0075 & 0.0291 & 0.0069 \\
 &  & CV & \bfseries 0.0064 & 0.0276 & \bfseries 0.0058 & 0.0062 & 0.0436 & \bfseries 0.0044 & \bfseries 0.0045 & 0.0307 & \bfseries 0.0037 \\
	\bottomrule
	\end{tabular}
	\label{tab:res-ale-f1}
\end{table}

\begin{table}[h]
    \centering
    \scriptsize
    \caption{Results for \ac{PD} on \textit{Feynman I.29.16} averaged over 100 grid points. Bold numbers indicate the minimum per metric-feature-model-size-combination.}
   \begin{tabular}{lll|rrr|rrr|rrr}
   \toprule
     &  & \textbf{Feature} & \multicolumn{3}{c}{$\mathbf{x_1}$} & \multicolumn{3}{c}{$\mathbf{\theta_1}$} & \multicolumn{3}{c}{$\mathbf{d_1}$} \\
     &  & \textbf{Metric} & MSE & Bias & Var & MSE & Bias & Var & MSE & Bias & Var \\
     \midrule
\multirow[c]{12}{*}{\rotatebox{90}{$n=1250$}} & \multirow[c]{3}{*}{GAM\_OF} & train & 0.2916 & 0.1357 & 0.2826 & 0.0660 & 0.0482 & 0.0659 & 0.0693 & 0.0497 & 0.0692 \\
 &  & val & 0.4493 & 0.1492 & 0.4417 & 0.0988 & 0.0617 & 0.0982 & 0.1011 & 0.0528 & 0.1017 \\
 &  & CV & \bfseries 0.1577 & \bfseries 0.0749 & \bfseries 0.1574 & \bfseries 0.0611 & \bfseries 0.0458 & \bfseries 0.0610 & \bfseries 0.0650 & \bfseries 0.0485 & \bfseries 0.0648 \\
	\cmidrule(){2-12}
 & \multirow[c]{3}{*}{GAM\_OT} & train & 0.0318 & 0.0358 & 0.0316 & 0.0199 & 0.0281 & 0.0197 & 0.0179 & 0.0194 & 0.0182 \\
 &  & val & 0.0430 & 0.0370 & 0.0431 & 0.0255 & 0.0310 & 0.0254 & 0.0215 & 0.0224 & 0.0217 \\
 &  & CV & \bfseries 0.0299 & \bfseries 0.0344 & \bfseries 0.0297 & \bfseries 0.0186 & \bfseries 0.0273 & \bfseries 0.0185 & \bfseries 0.0170 & \bfseries 0.0193 & \bfseries 0.0172 \\
	\cmidrule(){2-12}
 & \multirow[c]{3}{*}{XGB\_OF} & train & 0.0681 & 0.0623 & 0.0664 & 0.0032 & \bfseries 0.0209 & 0.0029 & 0.0030 & 0.0105 & 0.0030 \\
 &  & val & 0.0878 & 0.0715 & 0.0855 & 0.0049 & 0.0245 & 0.0044 & 0.0039 & 0.0178 & 0.0037 \\
 &  & CV & \bfseries 0.0429 & \bfseries 0.0408 & \bfseries 0.0426 & \bfseries 0.0021 & 0.0217 & \bfseries 0.0016 & \bfseries 0.0014 & \bfseries 0.0044 & \bfseries 0.0014 \\
	\cmidrule(){2-12}
 & \multirow[c]{3}{*}{XGB\_OT} & train & 0.0316 & \bfseries 0.0746 & 0.0270 & 0.0095 & 0.0229 & 0.0093 & 0.0074 & 0.0138 & 0.0074 \\
 &  & val & 0.0404 & 0.0813 & 0.0349 & 0.0117 & 0.0227 & 0.0116 & 0.0083 & 0.0162 & 0.0083 \\
 &  & CV & \bfseries 0.0297 & 0.0865 & \bfseries 0.0230 & \bfseries 0.0079 & \bfseries 0.0221 & \bfseries 0.0077 & \bfseries 0.0063 & \bfseries 0.0129 & \bfseries 0.0064 \\
 \midrule
\multirow[c]{12}{*}{\rotatebox{90}{$n=10000$}} & \multirow[c]{3}{*}{GAM\_OF} & train & 0.0146 & 0.0221 & 0.0146 & 0.0072 & 0.0214 & 0.0070 & 0.0071 & 0.0160 & 0.0071 \\
 &  & val & 0.0192 & 0.0253 & 0.0192 & 0.0090 & 0.0217 & 0.0088 & 0.0090 & 0.0184 & 0.0089 \\
 &  & CV & \bfseries 0.0142 & \bfseries 0.0219 & \bfseries 0.0142 & \bfseries 0.0070 & \bfseries 0.0212 & \bfseries 0.0068 & \bfseries 0.0069 & \bfseries 0.0156 & \bfseries 0.0068 \\
	\cmidrule(){2-12}
 & \multirow[c]{3}{*}{GAM\_OT} & train & 0.0061 & 0.0135 & 0.0061 & 0.0036 & 0.0179 & 0.0034 & 0.0033 & 0.0123 & 0.0033 \\
 &  & val & 0.0075 & 0.0166 & 0.0075 & 0.0040 & \bfseries 0.0156 & 0.0039 & 0.0039 & 0.0134 & 0.0039 \\
 &  & CV & \bfseries 0.0058 & \bfseries 0.0134 & \bfseries 0.0058 & \bfseries 0.0034 & 0.0178 & \bfseries 0.0032 & \bfseries 0.0032 & \bfseries 0.0121 & \bfseries 0.0032 \\
	\cmidrule(){2-12}
 & \multirow[c]{3}{*}{XGB\_OF} & train & 0.0103 & 0.0198 & 0.0103 & 0.0014 & 0.0293 & 0.0006 & 0.0007 & 0.0080 & 0.0006 \\
 &  & val & 0.0124 & 0.0207 & 0.0124 & 0.0017 & 0.0293 & 0.0008 & 0.0007 & 0.0078 & 0.0007 \\
 &  & CV & \bfseries 0.0066 & \bfseries 0.0156 & \bfseries 0.0065 & \bfseries 0.0012 & \bfseries 0.0290 & \bfseries 0.0004 & \bfseries 0.0004 & \bfseries 0.0069 & \bfseries 0.0004 \\
	\cmidrule(){2-12}
 & \multirow[c]{3}{*}{XGB\_OT} & train & 0.0057 & \bfseries 0.0181 & 0.0056 & 0.0028 & 0.0203 & 0.0024 & 0.0023 & 0.0098 & 0.0023 \\
 &  & val & 0.0072 & 0.0204 & 0.0070 & 0.0033 & \bfseries 0.0202 & 0.0030 & 0.0026 & 0.0106 & 0.0025 \\
 &  & CV & \bfseries 0.0045 & 0.0185 & \bfseries 0.0043 & \bfseries 0.0024 & 0.0204 & \bfseries 0.0020 & \bfseries 0.0018 & \bfseries 0.0093 & \bfseries 0.0018 \\
	\bottomrule
	\end{tabular}
	\label{tab:res-pd-fmn}
\end{table}

\begin{table}[h]
    \centering
    \scriptsize
    \caption{Results for \ac{ALE} on \textit{Feynman I.29.16} averaged over 100 grid points. Bold numbers indicate the minimum per metric-feature-model-size-combination.}
   \begin{tabular}{lll|rrr|rrr|rrr}
   \toprule
     &  & \textbf{Feature} & \multicolumn{3}{c}{$\mathbf{x_1}$} & \multicolumn{3}{c}{$\mathbf{\theta_1}$} & \multicolumn{3}{c}{$\mathbf{d_1}$} \\
     &  & \textbf{Metric} & MSE & Bias & Var & MSE & Bias & Var & MSE & Bias & Var \\
     \midrule
\multirow[c]{12}{*}{\rotatebox{90}{$n=1250$}} & \multirow[c]{3}{*}{GAM\_OF} & train & 0.3122 & \bfseries 0.1226 & 0.3075 & \bfseries 0.0701 & 0.0591 & \bfseries 0.0689 & \bfseries 0.0738 & \bfseries 0.0506 & \bfseries 0.0737 \\
 &  & val & 0.7270 & 0.2335 & 0.6956 & 0.2011 & 0.0703 & 0.2029 & 0.1729 & 0.0797 & 0.1723 \\
 &  & CV & \bfseries 0.2393 & 0.2227 & \bfseries 0.1963 & 0.0712 & \bfseries 0.0503 & 0.0711 & 0.0786 & 0.0556 & 0.0781 \\
	\cmidrule(){2-12}
 & \multirow[c]{3}{*}{GAM\_OT} & train & \bfseries 0.0337 & \bfseries 0.0405 & 0.0332 & 0.0214 & \bfseries 0.0400 & 0.0205 & 0.0179 & 0.0195 & 0.0181 \\
 &  & val & 0.0827 & 0.1494 & 0.0625 & 0.0322 & 0.0437 & 0.0314 & 0.0220 & 0.0230 & 0.0222 \\
 &  & CV & 0.0604 & 0.1697 & \bfseries 0.0327 & \bfseries 0.0185 & 0.0409 & \bfseries 0.0174 & \bfseries 0.0146 & \bfseries 0.0183 & \bfseries 0.0148 \\
	\cmidrule(){2-12}
 & \multirow[c]{3}{*}{XGB\_OF} & train & 0.5272 & 0.3080 & 0.4473 & 0.1359 & 0.0539 & 0.1376 & 0.1118 & 0.0677 & 0.1109 \\
 &  & val & 1.8846 & \bfseries 0.2021 & 1.9073 & 0.1266 & 0.0616 & 0.1270 & 0.0653 & \bfseries 0.0428 & 0.0656 \\
 &  & CV & \bfseries 0.3804 & 0.2820 & \bfseries 0.3113 & \bfseries 0.0381 & \bfseries 0.0520 & \bfseries 0.0366 & \bfseries 0.0338 & 0.0502 & \bfseries 0.0323 \\
	\cmidrule(){2-12}
 & \multirow[c]{3}{*}{XGB\_OT} & train & \bfseries 0.0362 & \bfseries 0.0421 & 0.0356 & 0.0186 & 0.0379 & 0.0177 & 0.0129 & 0.0176 & 0.0130 \\
 &  & val & 0.1037 & 0.1760 & 0.0752 & 0.0318 & \bfseries 0.0277 & 0.0321 & 0.0189 & 0.0240 & 0.0189 \\
 &  & CV & 0.0790 & 0.2141 & \bfseries 0.0344 & \bfseries 0.0120 & 0.0397 & \bfseries 0.0108 & \bfseries 0.0076 & \bfseries 0.0146 & \bfseries 0.0076 \\
 \midrule
\multirow[c]{12}{*}{\rotatebox{90}{$n=10000$}} & \multirow[c]{3}{*}{GAM\_OF} & train & 0.0149 & 0.0257 & 0.0147 & 0.0079 & 0.0304 & 0.0072 & 0.0072 & 0.0162 & 0.0072 \\
 &  & val & 0.0219 & 0.0294 & 0.0218 & 0.0114 & 0.0308 & 0.0108 & 0.0096 & 0.0190 & 0.0096 \\
 &  & CV & \bfseries 0.0148 & \bfseries 0.0255 & \bfseries 0.0147 & \bfseries 0.0077 & \bfseries 0.0303 & \bfseries 0.0070 & \bfseries 0.0070 & \bfseries 0.0156 & \bfseries 0.0070 \\
	\cmidrule(){2-12}
 & \multirow[c]{3}{*}{GAM\_OT} & train & 0.0063 & 0.0191 & 0.0062 & 0.0041 & 0.0279 & 0.0035 & 0.0033 & 0.0123 & 0.0033 \\
 &  & val & 0.0081 & 0.0230 & 0.0078 & 0.0052 & \bfseries 0.0268 & 0.0046 & 0.0040 & 0.0134 & 0.0039 \\
 &  & CV & \bfseries 0.0061 & \bfseries 0.0191 & \bfseries 0.0059 & \bfseries 0.0040 & 0.0278 & \bfseries 0.0033 & \bfseries 0.0032 & \bfseries 0.0120 & \bfseries 0.0032 \\
	\cmidrule(){2-12}
 & \multirow[c]{3}{*}{XGB\_OF} & train & 0.0470 & 0.0386 & 0.0471 & 0.0199 & 0.0541 & 0.0175 & 0.0105 & 0.0331 & 0.0098 \\
 &  & val & 0.2569 & 0.0855 & 0.2582 & 0.0197 & 0.0568 & 0.0171 & 0.0129 & 0.0196 & 0.0129 \\
 &  & CV & \bfseries 0.0358 & \bfseries 0.0354 & \bfseries 0.0357 & \bfseries 0.0059 & \bfseries 0.0459 & \bfseries 0.0039 & \bfseries 0.0037 & \bfseries 0.0178 & \bfseries 0.0035 \\
	\cmidrule(){2-12}
 & \multirow[c]{3}{*}{XGB\_OT} & train & 0.0069 & \bfseries 0.0154 & 0.0069 & 0.0045 & \bfseries 0.0241 & 0.0041 & 0.0029 & 0.0116 & 0.0029 \\
 &  & val & 0.0154 & 0.0212 & 0.0154 & 0.0087 & 0.0293 & 0.0081 & 0.0043 & 0.0132 & 0.0043 \\
 &  & CV & \bfseries 0.0058 & 0.0256 & \bfseries 0.0054 & \bfseries 0.0043 & 0.0293 & \bfseries 0.0035 & \bfseries 0.0022 & \bfseries 0.0098 & \bfseries 0.0022 \\
	\bottomrule
	\end{tabular}
	\label{tab:res-ale-fmn}
\end{table}

\begin{table}[h]
    \centering
    \scriptsize
    \caption{Decomposition results of total variance $\text{Var}_\text{Tot}$ into model $\text{Var}_\text{Mod}$ and estimation variance $\text{Var}_\text{Est}$ on \textit{SimpleNormalCorrelated} averaged over 100 grid points. Bold numbers indicate estimation strategy with minimal variance.}
   \begin{tabular}{lll|rrr|rrr|rrr}
   \toprule
     &  & \textbf{Feature} & \multicolumn{3}{c}{$\mathbf{x_1}$} & \multicolumn{3}{c}{$\mathbf{x_2}$} & \multicolumn{3}{c}{$\mathbf{x_3}$} \\
     &  & \textbf{Metric} & $\text{Var}_\text{Tot}$ & $\text{Var}_\text{Mod}$ & $\text{Var}_\text{Est}$ & $\text{Var}_\text{Tot}$ & $\text{Var}_\text{Mod}$ & $\text{Var}_\text{Est}$ & $\text{Var}_\text{Tot}$ & $\text{Var}_\text{Mod}$ & $\text{Var}_\text{Est}$ \\
    \midrule
\multirow[c]{6}{*}{\rotatebox{90}{PD}} & \multirow[c]{3}{*}{XGB\_OF} & train & 0.0817 & 0.0813 & 0.0004 & 0.2263 & 0.2259 & 0.0004 & 0.0010 & 0.0009 & 0.0001 \\
 &  & val & 0.0857 & 0.0837 & 0.0020 & 0.2416 & 0.2395 & 0.0021 & 0.0014 & 0.0011 & 0.0003 \\
 &  & CV & \bfseries 0.0505 & \bfseries 0.0501 & \bfseries 0.0004 & \bfseries 0.1291 & \bfseries 0.1287 & \bfseries 0.0004 & \bfseries 0.0007 & \bfseries 0.0006 & \bfseries 0.0001 \\
 \cmidrule(){2-12}
 & \multirow[c]{3}{*}{XGB\_OT} & train & 0.0113 & 0.0112 & \bfseries 0.0001 & 0.0141 & 0.0141 & \bfseries 0.0001 & 0.0014 & 0.0014 & 0.0000 \\
 &  & val & 0.0123 & 0.0119 & 0.0004 & 0.0155 & 0.0151 & 0.0004 & 0.0019 & 0.0019 & 0.0000 \\
 &  & CV & \bfseries 0.0090 & \bfseries 0.0089 & 0.0001 & \bfseries 0.0104 & \bfseries 0.0103 & 0.0001 & \bfseries 0.0013 & \bfseries 0.0013 & \bfseries 0.0000 \\
 \midrule
\multirow[c]{6}{*}{\rotatebox{90}{ALE}} & \multirow[c]{3}{*}{XGB\_OF} & train & \bfseries 0.1012 & \bfseries 0.0161 & \bfseries 0.0851 & 0.1011 & 0.0249 & \bfseries 0.0762 & 0.0741 & 0.0636 & \bfseries 0.0105 \\
 &  & val & 0.5281 & 0.0359 & 0.4923 & 0.6017 & 0.1743 & 0.4275 & 0.0864 & 0.0424 & 0.0439 \\
 &  & CV & 0.1520 & 0.0620 & 0.0900 & \bfseries 0.0994 & \bfseries 0.0110 & 0.0884 & \bfseries 0.0121 & \bfseries 0.0003 & 0.0118 \\
 \cmidrule(){2-12}
 & \multirow[c]{3}{*}{XGB\_OT} & train & 0.0088 & 0.0077 & \bfseries 0.0011 & 0.0119 & 0.0107 & \bfseries 0.0011 & 0.0066 & 0.0062 & \bfseries 0.0005 \\
 &  & val & 0.0263 & 0.0124 & 0.0140 & 0.0275 & 0.0089 & 0.0186 & 0.0033 & \bfseries 0.0003 & 0.0029 \\
 &  & CV & \bfseries 0.0077 & \bfseries 0.0049 & 0.0029 & \bfseries 0.0103 & \bfseries 0.0065 & 0.0038 & \bfseries 0.0020 & 0.0014 & 0.0007 \\
	\bottomrule
	\end{tabular}
	\label{tab:res-ablation-snc}
\end{table}

\begin{table}[tb]
    \centering
    \scriptsize
    \begin{threeparttable}
    \caption{Decomposition results of total variance $\text{Var}_\text{Tot}$ into model $\text{Var}_\text{Mod}$ and estimation variance $\text{Var}_\text{Est}$ on \textit{Feynman I.29.16} averaged over 100 grid points. Bold numbers indicate estimation strategy with minimal variance.}
   \begin{tabular}{lll|rrr|rrr|rrr}
   \toprule
     &  & \textbf{Feature} & \multicolumn{3}{c}{$\mathbf{x_1}$} & \multicolumn{3}{c}{$\mathbf{\theta_1}$} & \multicolumn{3}{c}{$\mathbf{d_1}$} \\
     &  & \textbf{Metric} & $\text{Var}_\text{Tot}$ & $\text{Var}_\text{Mod}$ & $\text{Var}_\text{Est}$ & $\text{Var}_\text{Tot}$ & $\text{Var}_\text{Mod}$ & $\text{Var}_\text{Est}$ & $\text{Var}_\text{Tot}$ & $\text{Var}_\text{Mod}$ & $\text{Var}_\text{Est}$ \\
    \midrule
\multirow[c]{6}{*}{\rotatebox{90}{PD}} & \multirow[c]{3}{*}{XGB\_OF} & train & 0.0664 & 0.0652 & 0.0013 & 0.0029 & 0.0027 & 0.0002 & 0.0030 & 0.0029 & 0.0002 \\
 &  & val & 0.0855 & 0.0794 & 0.0061 & 0.0044 & 0.0035 & 0.0009 & 0.0037 & 0.0029 & 0.0008 \\
 &  & CV & \bfseries 0.0426 & \bfseries 0.0414 & \bfseries 0.0012 & \bfseries 0.0016 & \bfseries 0.0015 & \bfseries 0.0002 & \bfseries 0.0014 & \bfseries 0.0013 & \bfseries 0.0001 \\
	\cmidrule(){2-12}
 & \multirow[c]{3}{*}{XGB\_OT} & train & 0.0270 & 0.0266 & \bfseries 0.0004 & 0.0093 & 0.0091 & 0.0002 & 0.0074 & 0.0074 & 0.0000 \\
 &  & val & 0.0349 & 0.0334 & 0.0015 & 0.0116 & 0.0108 & 0.0007 & 0.0083 & 0.0081 & 0.0001 \\
 &  & CV & \bfseries 0.0230 & \bfseries 0.0226 & 0.0004 & \bfseries 0.0077 & \bfseries 0.0076 & \bfseries 0.0002 & \bfseries 0.0064 & \bfseries 0.0064 & \bfseries 0.0000 \\
 \midrule
\multirow[c]{6}{*}{\rotatebox{90}{ALE}} & \multirow[c]{3}{*}{XGB\_OF} & train & 0.4473 & 0.0498 & \bfseries 0.3975 & 0.1376 & 0.1135 & \bfseries 0.0240 & 0.1109 & 0.0864 & \bfseries 0.0245 \\
 &  & val & 1.9073 & \multicolumn{1}{c}{-\tnote{*}} & 2.3566 & 0.1270 & \multicolumn{1}{c}{-\tnote{*}} & 0.1507 & 0.0656 & \multicolumn{1}{c}{-\tnote{*}} & 0.1164 \\
 &  & CV & \bfseries 0.3113 & \multicolumn{1}{c}{-\tnote{*}} & 0.4167 & \bfseries 0.0366 & 0.0099 & 0.0267 & \bfseries 0.0323 & 0.0071 & 0.0252 \\
	\cmidrule(){2-12}
 & \multirow[c]{3}{*}{XGB\_OT} & train & 0.0356 & 0.0302 & \bfseries 0.0054 & 0.0177 & 0.0141 & \bfseries 0.0037 & 0.0130 & 0.0112 & \bfseries 0.0017 \\
 &  & val & 0.0752 & \bfseries 0.0186 & 0.0565 & 0.0321 & 0.0096 & 0.0226 & 0.0189 & 0.0076 & 0.0113 \\
 &  & CV & \bfseries 0.0344 & 0.0227 & 0.0116 & \bfseries 0.0108 & \bfseries 0.0065 & 0.0043 & \bfseries 0.0076 & \bfseries 0.0053 & 0.0024 \\
	\bottomrule
	\end{tabular}
	\label{tab:res-ablation-fmn}
	\begin{tablenotes}
        \item[*] 
        \scriptsize Due to instabilities in variance estimation for ALE, likely caused by unstable bin assignments, the estimated $\text{Var}_\text{Est}$ sometimes exceeds the estimated $\text{Var}_\text{Tot}$. We omit $\text{Var}_\text{Mod}$ in these cases.
    \end{tablenotes}
    \end{threeparttable}
\end{table}

\begin{figure*}[h]
    \centering
    \begin{subfigure}[b]{0.49\textwidth}
        \centering
        \includegraphics[width=\textwidth]{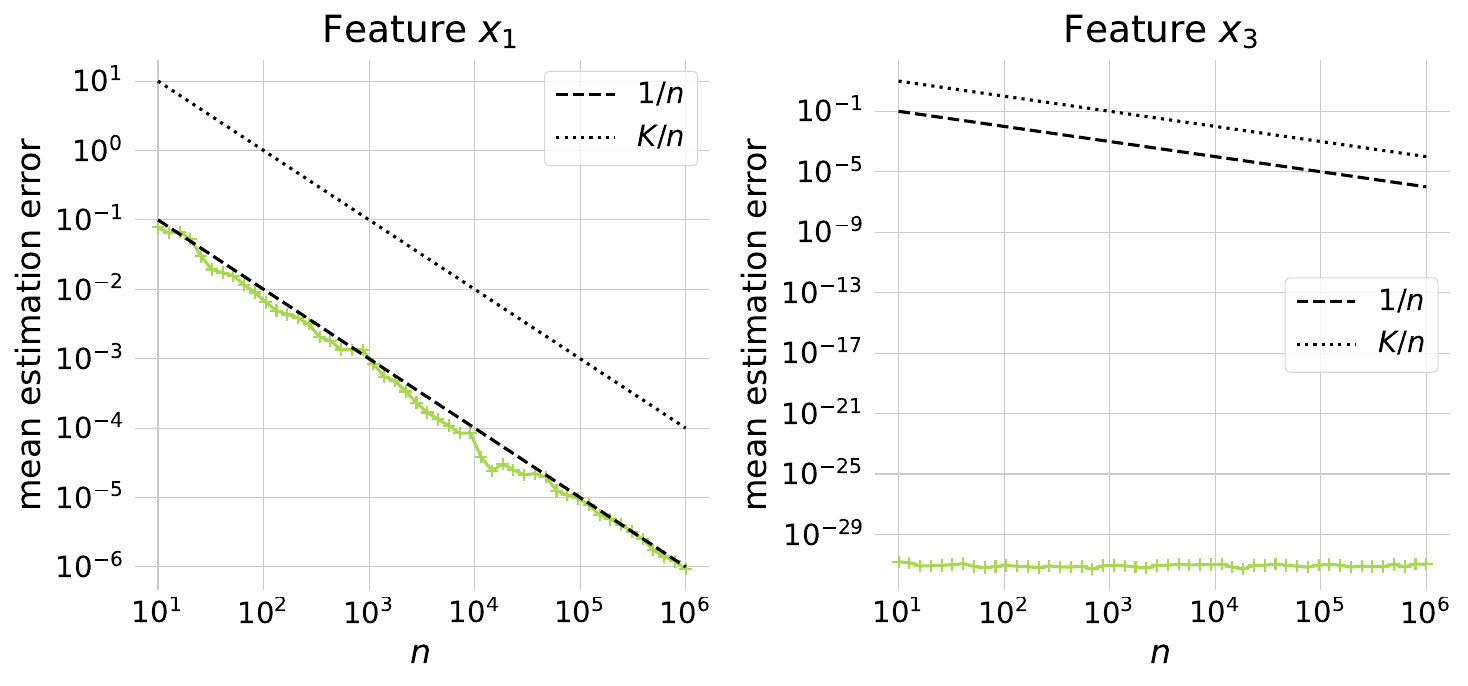}
        \caption{Mean estimation error \ac{PD}}  
        \label{fig:estim-error-pd-snc}
    \end{subfigure}
    \begin{subfigure}[b]{0.49\textwidth}
        \centering
        \includegraphics[width=\textwidth]{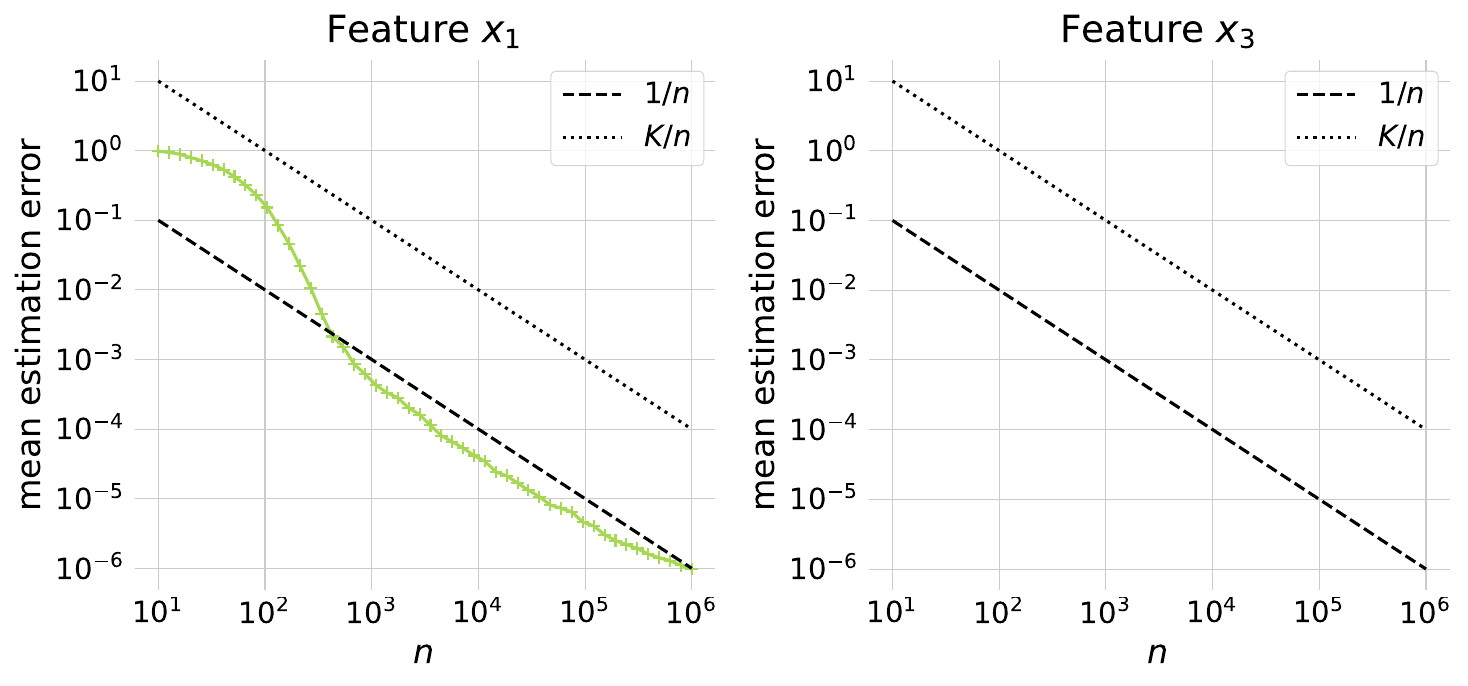}
        \caption{Mean estimation error \ac{ALE}}
        \label{fig:estim-error-ale-snc}
    \end{subfigure}
    \caption{Mean estimation errors on \textit{SimpleNormalCorrelated} for $X_1$ and $X_3$. For each sample size $\nsamples$, variances are averaged over all grid points. Axes are log-scale.}
    \label{fig:estim-error-snc}
\end{figure*}

\end{document}